%% file: main.tex
\documentclass{article}

\PassOptionsToPackage{numbers, compress}{natbib}

\usepackage[utf8]{inputenc} 
\usepackage[T1]{fontenc}    
\usepackage{hyperref}       
\usepackage{url}            
\usepackage{booktabs}       
\usepackage{amsfonts}       
\usepackage{nicefrac}       
\usepackage{microtype}      
\usepackage[hybrid,pipeTables=true]{markdown}

\usepackage[nomargin,inline,marginclue,draft]{fixme}
\usepackage{enumitem}
\usepackage{graphicx}
\usepackage{amsthm, amssymb}
\usepackage{mathtools}
\usepackage{bbold}
\usepackage{natbib}
\usepackage[title]{appendix}
\usepackage{geometry}
\hypersetup{colorlinks,citecolor=blue,urlcolor=blue,linkcolor=blue}
\usepackage{stackengine,scalerel}
\usepackage{multirow}

\usepackage{comment}
\usepackage{xcolor}
\usepackage{tabularx} 
\usepackage{adjustbox}

\newlength{\defbaselineskip}
\newcommand{\beginsupplement}{%
        \setcounter{table}{0}
        \renewcommand{\thetable}{S\arabic{table}}%
        \setcounter{figure}{0}
        \renewcommand{\thefigure}{S\arabic{figure}}%
     }

\newcommand{\totalclinicians}{15}
\newcommand{\totalspecialites}{7}
\newcommand{\dataset}{\textsc{MedAlign}}

\newcommand{\numinstruct}{983}  
\newcommand{\totalinstructions}{1314}
\newcommand{\instructionsinternalmedicine}{492}
\newcommand{\instructionsneurology}{320}
\newcommand{\instructionsradiology}{402}
\newcommand{\instructionscardiology}{71}
\newcommand{\instructionsoncology}{14}
\newcommand{\instructionssurgery}{12}
\newcommand{\instructionsprimarycare}{3}

\newcommand{\patientsingeneral}{455}
\newcommand{\patientswithuniquecharacteristics}{859}

\newcommand{\patientsingeneralfiltered}{407}

\newcommand{\ehrsamples}{77200}

\newcommand{\numannotated}{303}

\newcommand{\numehr}{276}
\newcommand{\numvisits}{27,150 }

\newcommand{\numcategories}{6}
\newcommand{\numsubcategories}{20}

\newlist{customenumC}{enumerate}{1}
\setlist[customenumC]{label=C\arabic*.}

\newcolumntype{R}[1]{>{\raggedright\arraybackslash}p{#1}}
\newcolumntype{C}[1]{>{\centering\arraybackslash}p{#1}}

\title{\dataset: A Clinician-Generated Dataset for Instruction Following with Electronic Medical Records}

\usepackage{authblk}
\author[,1,2]{Scott L. Fleming\thanks{Equal contributions. Corresponding author: \{scottyf, lozanoe\}@stanford.edu.}}
\author[$\ast$,1]{Alejandro Lozano}
\author[$\ast$,3,4]{William J. Haberkorn}
\author[$\ast$,5]{Jenelle A. Jindal}
\author[$\ast$,6,7,8]{Eduardo Reis}
\author[9]{Rahul Thapa}
\author[10]{Louis Blankemeier}
\author[11,12]{Julian Z. Genkins}
\author[2]{\\Ethan Steinberg}
\author[13]{Ashwin Nayak}
\author[5]{Birju Patel}
\author[14,15]{Chia-Chun Chiang}
\author[5,13]{Alison Callahan}
\author[5]{\\Zepeng Huo}
\author[6]{Sergios Gatidis}
\author[6]{Scott Adams}
\author[13]{Oluseyi Fayanju}
\author[13]{Shreya J. Shah}
\author[1,16]{\\Thomas Savage}
\author[5,17]{Ethan Goh}
\author[1,6,15]{Akshay S. Chaudhari}
\author[1,3,4]{Nima Aghaeepour}
\author[13,15]{Christopher Sharp}
\author[9,13]{Michael A. Pfeffer}
\author[2,15]{Percy Liang}  
\author[5,15,16,17]{Jonathan H. Chen}
\author[4]{\\Keith E. Morse}
\author[,2,15]{Emma P. Brunskill\thanks{Equal leadership.}}
\author[$\dagger$,5]{Jason A. Fries}
\author[$\dagger$,9,13,15,17]{Nigam H. Shah}

\affil[1]{Department of Biomedical Data Science, Stanford School of Medicine, Stanford, CA, USA}
\affil[2]{Department of Computer Science, Stanford School of Engineering, Stanford, CA, USA}
\affil[3]{Department of Anesthesiology, Peri-operative, and Pain Medicine, Stanford School of Medicine, Stanford, CA, USA}
\affil[4]{Department of Pediatrics, Stanford School of Medicine, Stanford, CA, USA}
\affil[5]{Stanford Center for Biomedical Informatics Research, Stanford University, Stanford, CA, USA}
\affil[6]{Department of Radiology, Stanford School of Medicine, Stanford, CA, USA}
\affil[7]{Center for Artificial Intelligence in Medicine and Imaging (AIMI), Stanford University, Stanford, CA, USA}
\affil[8]{Hospital Israelita Albert Einstein, Sao Paulo, SP, Brazil}
\affil[9]{Technology and Digital Solutions, Stanford Health Care, Palo Alto, CA, USA}
\affil[10]{Department of Electrical Enginering, Stanford School of Engineering, Stanford, CA}
\affil[11]{Department of Medicine, Vanderbilt University School of Medicine, Nashville, TN, USA}
\affil[12]{Department of Biomedical Informatics, Vanderbilt University Medical Center, Nashville, TN, USA}
\affil[13]{Department of Medicine, Stanford School of Medicine, Stanford, CA, USA}
\affil[14]{Department of Neurology, Mayo Clinic, Rochester, MN, USA}
\affil[15]{Human-Centered Artificial Intelligence Institute, Stanford University, Stanford, CA, USA}
\affil[16]{Division of Hospital Medicine, Stanford University, Stanford, CA, USA}
\affil[17]{Clinical Excellence Research Center, Stanford School of Medicine, Stanford, CA, USA}

\begin{document}

\maketitle


\begin{abstract}
\input{0_abstract}
\end{abstract}

\newpage

\input{1_introduction}

\input{3_dataset}

\input{4_results}


\input{6_discussion}

\bibliographystyle{abbrvnat}
\bibliography{references}

\newpage

\begin{appendices}
\beginsupplement
\appendix
\section*{Appendices}
\input{99_supplement}

\end{appendices}

\end{document}

%% file: 0_abstract.tex
The ability of large language models (LLMs) to follow natural language instructions with human-level fluency suggests many opportunities in healthcare to reduce administrative burden and improve quality of care.
However, evaluating LLMs on realistic text generation tasks for healthcare remains challenging. 
Existing question answering datasets for electronic health record (EHR) data fail to capture the complexity of information needs and documentation burdens experienced by clinicians. 
To address these challenges, we introduce \dataset, a benchmark dataset of \numinstruct\ natural language instructions for EHR data. 
\dataset\ is curated by \totalclinicians\ clinicians (\totalspecialites\ specialities), includes clinician-written reference responses for \numannotated\ instructions, and provides  \numehr\ longitudinal EHRs for grounding instruction-response pairs.
We used \dataset\ to evaluate 6 general domain LLMs, having clinicians rank the accuracy and quality of each LLM response. 
We found high error rates, ranging from 35\% (GPT-4) to 68\% (MPT-7B-Instruct), and 8.3\% drop in accuracy moving from 32k to 2k context lengths for GPT-4.
Finally, we report correlations between clinician rankings and automated natural language generation metrics as a way to rank LLMs without human review. 
\dataset\ is provided under a research data use agreement\footnote{https://medalign.stanford.edu} to enable LLM evaluations on tasks aligned with clinician needs and preferences.

%% file: 1_introduction.tex
\section{Introduction}


\begin{figure*}[!ht]
\includegraphics[width=1.0\textwidth]{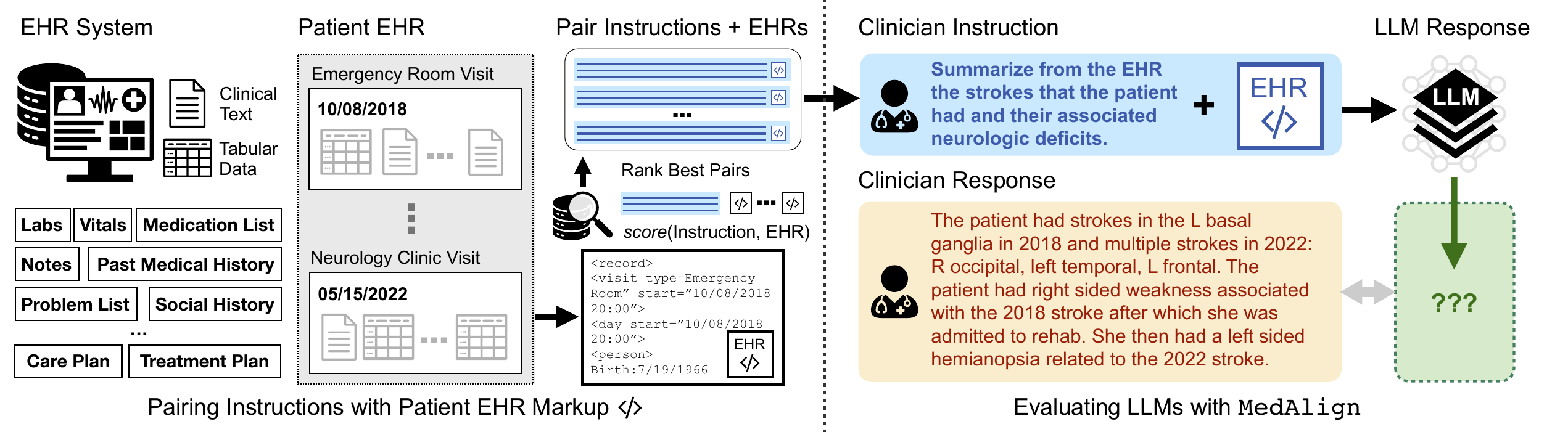}
\centering
\caption{In \dataset, patient EHRs are transformed into XML markup (example provided in Figure \ref{fig:xml_markup_example}) and paired with clinician-generated instructions using a retrieval-based (BM25) scoring metric. The resulting set of instruction + EHR pairs is then reviewed by clinicians to write gold responses, which are used to evaluate EHR instruction following in large language models  }
\label{fig:medalign}
\end{figure*}


Large language models (LLMs) have revolutionized natural language processing in tasks such as reading comprehension, reasoning, and language generation \cite{zhao2023survey}, prompting researchers to explore applications in healthcare \cite{thirunavukarasu2023large}. 
Recent LLMs like MedPalm \cite{singhal2023large} and GPT-4 \cite{nori2023capabilities} have demonstrated expert-level performance on medical question-answering benchmarks including MedQA \cite{jin2021disease}, 
MMLU \cite{hendrycks2020measuring}, 
and the USMLE \cite{kung2023performance}. However, these benchmarks employ multiple-choice, exam-style evaluations where question stems summarize key information and a single answer choice is best. It is not known if performance on these tasks will translate when a model is deployed in the complex clinical environments. 

To be useful, LLMs need to perform well on the specific information-related tasks that clinicians currently complete themselves while caring for patients. These tasks are a significant burden on clinicians, who spend 45\% of their day interacting with computers instead of patients \cite{toscano2020physicians} and 10 hours a week generating documentation \cite{gaffney2022medical}, in part contributing to professional burnout \cite{muhiyaddin2021electronic}. Examples of these tasks include summarizing a patient's asthma treatment history from different specialists the patient has visited, generating a differential diagnosis based on partially resulted laboratory data, or searching through the clinical notes for mentions of a patient's family support system in order to create the best plan for the patient's hospital discharge (see Table \ref{tab:category_analysis}). 
Such tasks could be passed as  instructions to an LLM in the form of questions or imperatives (e.g., ``Write a discharge summary'') grounded in a patient's Electronic Health Record (EHR, an electronic representation of a patient's medical history).
However, despite the excitement about LLMs to transform the practice of medicine, evaluations to date have not authentically represented the variety of tasks and idiosyncrasies of EHR data that clinicians face in the real world. 

Given the recent emergence of instruction-following capabilities in LLMs \cite{wei2022finetuned}, there is potential for LLMs to ameliorate such administrative burden. Hand-curated exemplars of instructions and responses have been critical to improve performance of models \cite{chung2022scaling}, especially on clinical reasoning and knowledge recall tasks in the healthcare domain \cite{singhal2023large}. Thus, a high quality dataset of instruction-EHR-response tuples that represents the breadth of clinical tasks is essential not only as a shared benchmark, but potentially to accelerate the training of specialized LLMs for healthcare \cite{shah2023creation}.

However, building such a dataset requires an extraordinary effort from a multidisciplinary collaboration. In particular, generating an instruction-following benchmark dataset with representative EHR-based tasks and expert responses is challenging due to the substantial cost and logistical complexity of clinician review. There is a need for an EHR dataset that (1) contains a diverse set of questions and instructions generated by practicing clinicians; (2) pairs these queries with EHRs from both inpatient and ambulatory care settings; (3) leverages both structured and unstructured data from the longitudinal EHR; and (4) is available to the broader academic community.

\input{tables/comparison_literature_short}


In light of these challenges and opportunities, we present three contributions:

\begin{enumerate}

\item  \textbf{\dataset\ Dataset:} We introduce a benchmark dataset called \dataset\, consisting of \numinstruct\ questions and instructions submitted by 
\totalclinicians\ practicing clinicians spanning \totalspecialites\ medical specialties. For \numannotated\ of these instructions, we provide a clinician-written reference answer and paired EHR for grounding prompts. Each clinician evaluated and ranked outputs from 6 different LLMs on these \numannotated\ instructions and wrote ``gold standard'' answers. To our knowledge, \dataset\ is the first dataset of EHR-based instruction-answer pairs (including question \emph{and} imperative instructions) written by clinicians, with clinician evaluations of LLM-generated outputs. Table \ref{tab:prior_work} summarizes \dataset\ and its distinction from existing datasets for clinical information needs.

\item  \textbf{Automated Instruction-EHR Matching:} We demonstrate the feasibility of a simple retrieval-based approach to pair an instruction with a relevant patient EHR. By isolating the process of instruction solicitation, we were able to scale and diversify the set of clinicians who submitted instructions. Furthermore, we show that our process for matching instructions to relevant EHRs produces a relevant pairing 74\% of the time --- at least twice as frequently as randomly pairing instructions to EHRs.

\item  \textbf{Automated Evaluation of LLM Responses:}  We analyze the correlation between clinician rankings and automated natural language generation (NLG) metrics as a way to scalably reproduce such analyses, reducing future needs for clinicians to label and rank LLM responses. 

\end{enumerate}


\section{Background and Related Work}

The volume of patient care data is growing exponentially, with a compound annual growth rate approaching 36\% \cite{culbertson2021skyrocketing}. 
Utilizing LLMs to more efficiently interact with patient data holds great potential to help clinicians manage increasingly complicated information needs and circumvent low-usability EHR interfaces \cite{melnick2020association}. However, evaluation of LLMs to improve meaningful outcomes like clinician burnout or patient health has been inadequately studied, mainly due to benchmark datasets which do not represent true clinician needs \cite{henry20202018}, narrowly focus on a specific medical specialty or subset of EHR data \cite{lehman-etal-2022-learning}, and/or are overly simplistic due to templated question construction \cite{pampari-etal-2018-emrqa, yue-etal-2020-clinical}. These works highlight the challenges in collecting high-quality clinician-generated questions and answers; we consider each in turn.

Questions and instructions in an EHR-based benchmark dataset should be paired with relevant patient EHRs. In order to ensure relevancy, prior works have provided clinicians with specific patient EHRs and asked them to generate questions based on those patients' data \cite{lehman-etal-2022-learning}. Unfortunately, requiring EHRs as context for question generation limits scalability, as medical institutions restrict access to patient data to preserve patient privacy. \citet{pampari-etal-2018-emrqa} attempted to overcome these scalability issues by generating questions via a template-based approach, but this led to issues with question quality and diversity \cite{yue-etal-2020-clinical}. Our method of soliciting clinician-generated instructions without a specific patient's EHR as context overcomes these scaling issues, albeit at the cost of potentially less relevant instruction-to-EHR pairings (we discuss our approach to addressing this problem in the Dataset Curation section). 

Beyond generating questions, generating expert answers at scale is also prohibitively difficult. 
Reviewing an EHR to answer patient-specific queries can take 30+ minutes for a single patient \cite{siems2020structured}.
This excludes any time required to generate a response to the query. 
Prior works have attempted to overcome the bottleneck of generating responses by extracting answers verbatim from individual clinical notes or discharge summaries \cite{soni2022radqa,oliveira2021experiments,fan2019annotating}. However, many clinical tasks require synthesizing information from both structured data and multiple free-text documents to arrive at an adequate response, an aspect not explored in existing EHR QA datasets.
In such cases, answers extracted from a single note in the patient's record may not be an adequate; free-text text generation is required. While there is at least one example of an EHR-based question answering dataset in the literature that includes both structured and unstructured data \cite{raghavan2018annotating}, it neither contains free-text responses nor is publicly available. Finally, all of the aforementioned datasets focus on simple question answering (i.e., providing concise, factoid-style answers) rather than general instruction following, which often requires executing a series of complex directives and commands to accomplish tasks. To the best of our knowledge, there does not exist \emph{any} EHR-based benchmark dataset that incorporates instruction following.

The significant costs of clinician review present barriers not only for \emph{de novo} dataset generation, but also for reliable evaluation of new methods on existing datasets. Automated metrics for evaluating Natural Language Generation (NLG) systems have shown moderate to high correlation with human judgments on tasks like machine translation \cite{freitag2022results}, but it is unclear whether these findings extend to other domains and tasks. While there is precedent \cite{lehman-etal-2022-learning} for \emph{applying} automated metrics like BLEU \cite{papineni2002bleu}, ROUGE-L \cite{lin2004rouge}, METEOR \cite{banerjee2005meteor}, and BERTScore \cite{zhang2019bertscore} to NLG tasks in the clinical domain, there is comparatively very little work assessing correspondence between these metrics and human judgment on clinical NLG tasks. Thus not only do we have a poor understanding of how LLMs perform on EHR-based instruction-following tasks, but also we do not know whether it is possible to reliably automate such evaluations. Automation could substantially reduce the ``barrier to entry'' for research teams with limited resources.

%% file: tables/comparison_literature_short.tex
\begin{table*}[t]
\caption{Comparison of our work, \dataset, to existing EHR QA datasets.}
\centering
\normalsize
\begin{adjustbox}{center}
\begin{tabular}{C{3cm}C{1.5cm}C{1.5cm}C{1.8cm}C{1.3cm}C{3cm}C{3.5cm}C{2cm}}
\toprule
\textbf{Dataset} & \textbf{Questions} & \textbf{Documents} & \textbf{Patients} & \textbf{Specialties} & \textbf{Labeler} & \textbf{Source} \\
\midrule
\citet{raghavan2018annotating} & 5696 & 71 & 71 & - & Medical Students & Clinical Note \\
\citet{pampari-etal-2018-emrqa} & 73111 & 303 & 303 & - & Programmatic & Discharge Summary \\
\citet{fan2019annotating} & 245 & 138 & - & 1 & Author & Discharge Summary \\
\citet{yue2021cliniqg4qa} & 1287 & 36 & - & - & Medical Experts & Clinical Note \\
\citet{soni2022radqa} & 3074 & 1009 & 100 & 1 & Clinicians & Radiology Note \\
\midrule
\dataset\ (Ours) & 983 & 37264 & 276 & \totalspecialites & Clinicians & EHR \\    
\bottomrule
\end{tabular}
\end{adjustbox}
\label{tab:prior_work}
\end{table*}

%% file: 3_dataset.tex
\section{Dataset Curation Process}

\paragraph{Electronic Health Records (EHRs)} 
EHR systems are software for managing patient medical record data. 
From a clinician's view, a patient EHR is accessed via a graphical user interface that provides access to data elements associated with medical care, e.g., medication lists and treatment plans.
These data are stored as a collection of timestamped structured (tabular) and unstructured (text) events, which when ordered by time form a patient's longitudinal EHR timeline.    
Our EHR data is represented using the OMOP CDM \cite{voss2015feasibility}, a standardized schema for exchanging medical data, translated into a single, XML markup document per record (example provided in Figure \ref{fig:xml_markup_example}) to enable simple data exploration via an XML viewer.  
Figure \ref{fig:medalign} outlines the workflow for building \dataset\ including (1) pairing clinician-generated instructions with patient EHR markup, and (2) evaluating language model responses against gold responses written by clinicians.

\paragraph{Collection Protocol}  
Reviewing patient medical data requires adhering to strict security protocols to protect patient privacy and prevent protected health information (PHI) leaks. 
This motivated our 3-stage curation process: (1) online instruction collection from clinicians; (2) instruction-EHR matching; and (3) response generation.
Note we deliberately decouple instruction collection from response generation.
This enables sampling a larger set of instructions from a more diverse set of clinician specialities while minimizing exposure to patient data. 
However, this approach requires defining a matching function to pair instructions with relevant patient EHRs, a process which may generate errors due to irrelevant instruction-EHR pairings.
We discuss the performance of a retrieval-based matching system below.

\paragraph{Stage 1: Collecting Instructions} 
Clinicians were recruited in our academic medical center via email.
Through the use of an online form, clinicians were asked to submit instructions as posed to a hypothetical AI assistant designed to facilitate EHR-based tasks.
Participants were instructed to envision a clinical vignette typical of their daily practice and to formulate an instruction that the AI could perform to make their work easier, faster, and less stressful. 
For each instruction, participants were asked to provide metadata to assist in matching the instruction to a patient, including pertinent clinical characteristics and the clinical context where the instruction could be used, e.g., ``when deciding whether to use contrast in a CT scan''. 
See Appendix \ref{app:instruction-collection-form} for all collected fields.

\input{tables/instruction_categories}


\paragraph{Stage 2: Instruction-EHR matching}

All submitted instructions include  metadata information on their intended clinical context and target patient population.
We used instructions tagged ``applicable
to patients generally'' to maximize their relevance in EHR matching. 
We evaluated two methods for matching instructions with EHRs: (1) a simple baseline based on uniform random sampling; and (2) a retrieval-based method using BM25Okapi \cite{trotman2014improvements}. 

For the retrieval approach, we concatenated every instruction with its corresponding  patient characteristics and clinical context to construct a search query.
We used this query to retrieve the 5 most relevant EHRs within a randomly selected subsample of \ehrsamples\  patients from our hospital database. 
This same subsample was used to match patients for our baseline uniform random sample.
After matching, the authors conducted a manual review to assess binary relevance of all generated instruction-EHR pairs.

\paragraph{Stage 3: Instruction Response Generation}

For this stage, clinicians were tasked with reviewing the instruction and associated EHR data, then writing a response to that instruction. Whenever feasible, instructions were assigned to clinicians within the same specialty as the original submitter but not the original submitter themselves. In cases where this was not possible, the instruction was randomly assigned to a clinician, in any specialty, that did not submit the instruction. Clinicians were asked whether the instruction could be feasibly applied to the patient in the EHR (e.g., not asking about smoking history in an infant) and if the EHR contained all necessary information to answer the instruction. They then manually generated an expert response to the instruction. This response was intended to be brief and clinically relevant, drawing on any information available in the supplied EHR, as well as any appropriate external references. 
The most recent timestamp in the EHR was designated as the ``time anchor'', meaning the response was written as if the instruction had been posed at that point in time. 



\section{Dataset Description}

\paragraph{Instructions Collected} 
A total of \totalclinicians\ clinicians submitted instructions during the data collection process. 
These medical practitioners represented 7 distinct specialties, which included Internal Medicine (\instructionsinternalmedicine\ instructions submitted), Neurology (\instructionsneurology), Radiology (\instructionsradiology), Cardiology (\instructionscardiology), Oncology (\instructionsoncology), Surgery (\instructionssurgery), and Primary Care (\instructionsprimarycare). 
Clinicians provided a varying number of instructions ranging from 1 to 278 with a mean of 87 instructions per clinician (see Figure \ref{fig:breakdown_clincian_submitters}). From the \totalinstructions\ instructions collected, \patientsingeneral\ were marked as applicable to patients generally and \patientswithuniquecharacteristics\ were relevant only to patients with specific clinical characteristics. We removed  near-identical instructions (defined by a  ROUGE-L similarity above  0.7), yielding \numinstruct\ instructions of which \patientsingeneralfiltered\ were marked as applicable to patients generally.

\paragraph{Instruction-EHR Matches} 
Based on evaluation by the authors, for 240 (59\%) of the instructions applicable to ``patients in general'' the first record retrieved by BM25 was relevant. For 303 instructions (74\%), at least one of the top 5 EHRs returned by BM25 was relevant. In contrast, only 38\% of EHRs retrieved via uniform random sampling were deemed relevant.

\paragraph{Instruction Taxonomy} 
To better understand higher-level themes within the instructions submitted, a practicing clinician developed a taxonomy of instructions. 
This taxonomy, described in detail in Table \ref{tab:subcategory_analysis}, includes 6 categories spanning 20 subcategories. We summarize the distribution of instruction categories across the set of all instructions submitted and those that received responses from a clinician in Table \ref{tab:category_analysis}. 

%% file: tables/instruction_categories.tex
\begin{table}[tb]
\centering
\normalsize
\renewcommand{\arraystretch}{1.25} 
\caption{\dataset\ instruction categories and example instructions.}
\begin{adjustbox}{center}
\begin{tabular}{R{3.8cm}R{8.5cm}R{0.8cm}R{0.8cm}}
\toprule
\textbf{Category} & \textbf{Example Instruction} & \textbf{Gold} & \textbf{All} \\
\midrule
Retrieve \& Summarize & Summarize the most recent annual physical with the PCP & 223 & 667 \\
Care Planning & Summarize the asthma care plan for this patient including relevant diagnostic testing, exacerbation history, and treatments & 22 & 136 \\
Calculation \& Scoring & Identify the risk of stroke in the next 7 days for this TIA patient & 13 & 70 \\
Diagnosis Support & Based on the information I've included under HPI, what is a reasonable differential diagnosis? & 4 & 33 \\
Translation & I have a patient that speaks only French. Please translate these FDG-PET exam preparation instructions for her & 0 & 2 \\
Other & What patients on my service should be prioritized for discharge today? & 41 & 75 \\
\midrule
Total & & 303 & 983 \\
\bottomrule
\end{tabular}
\end{adjustbox}
\label{tab:category_analysis}
\end{table}

%% file: 4_results.tex
\input{tables/human_eval_rank_acc}

\section{Benchmarking LLM Performance}

\begin{figure}[ht]
\includegraphics[width=0.45\textwidth]{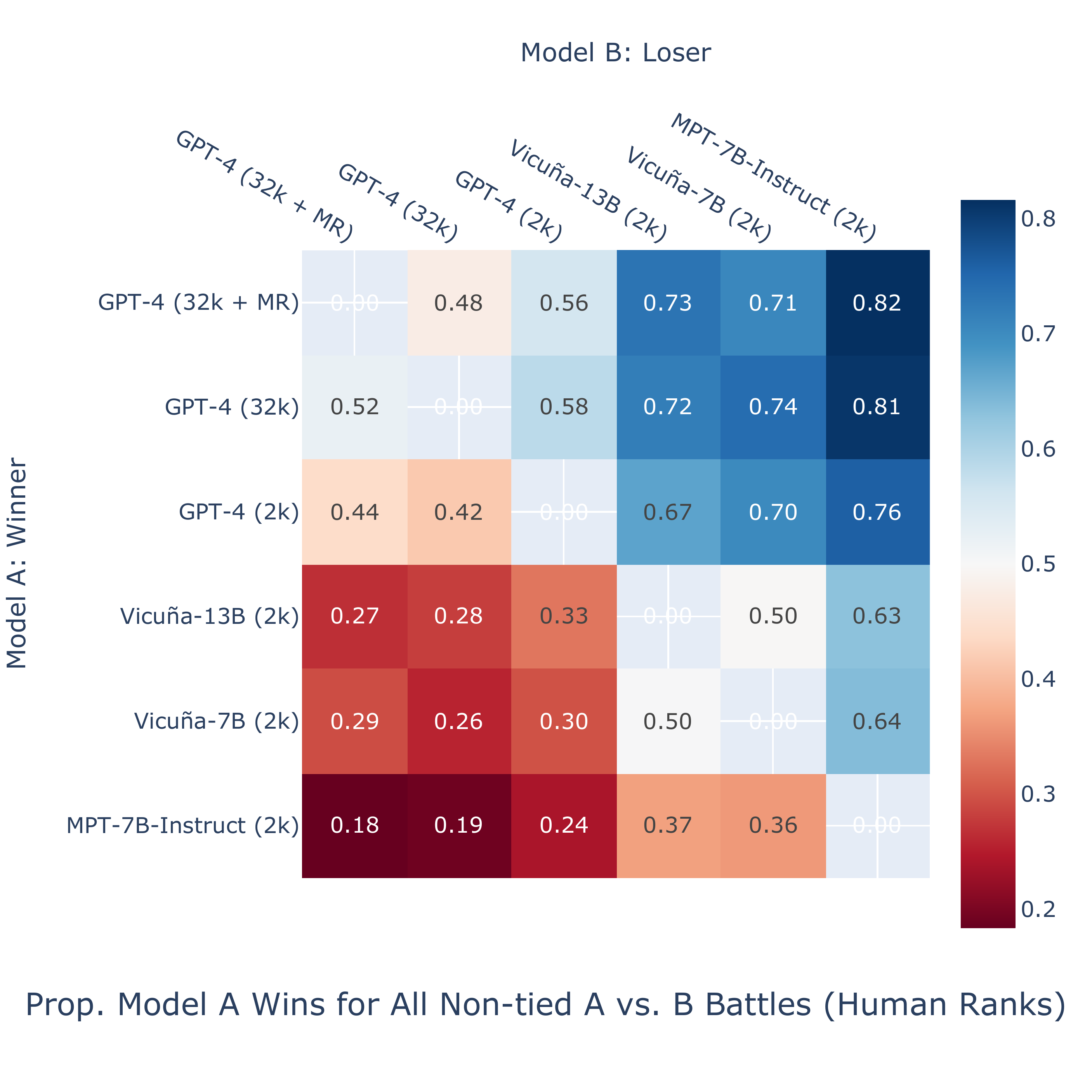}
\includegraphics[width=0.45\textwidth]{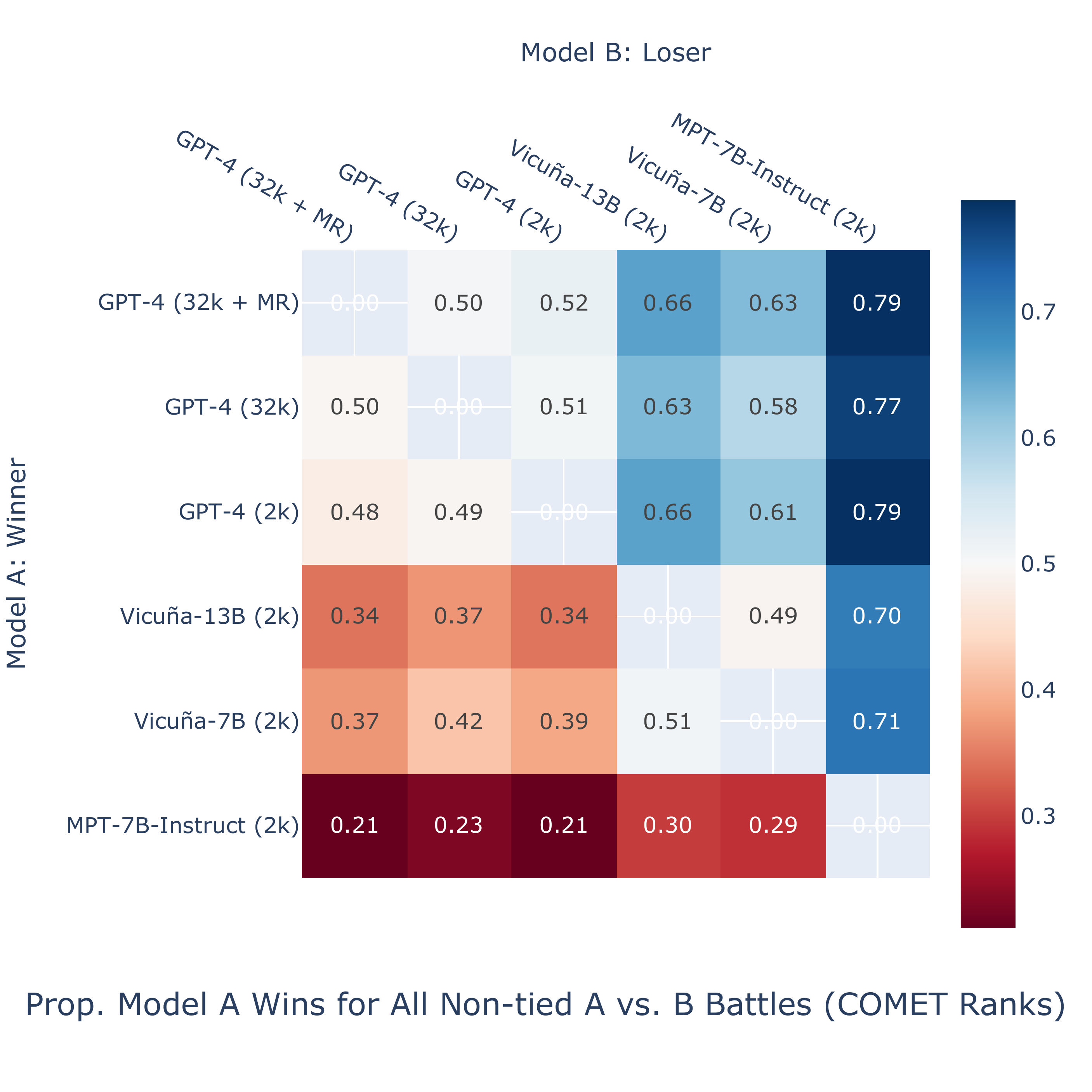}
\centering
\caption{(Left) Head-to-head comparison of model performance based on human ranks. The number in row $i$, column $j$ indicates the proportion of instructions for which the response generated by the model in row $i$ was strictly preferred over the model in column $j$. (Right) Head-to-head evaluation of model performance using COMET Ranks. Represents the same matrix structure and interpretation as on the left, but using rankings derived from COMET, an automated metric, rather than clinician-generated rankings. Model win rates using COMET follow a similar pattern as to model win rates using human rankings.}
\label{fig:head_to_head_win_rate_human}
\end{figure}

\paragraph{LLM Selection}
We evaluated six distinct LLMs, chosen to capture both state-of-the-art, closed-source LLM capabilities available to consumers via an API as well as smaller, open-source and user-modifiable LLMs with more lenient commercial licensing (e.g., MosaicML's MPT-7B-Instruct model). Additionally, we designed our experiments to directly evaluate the impact of model parameters and context length.

For a state-of-the-art LLM, we selected GPT-4 (through Microsoft's Azure OpenAI HIPAA compliant gpt-4-32k-0301 API) due to its state-of-the-art performance on various medical tasks, its long 32k context length, and its availability to researchers and clinics. However, despite this context length, it proved insufficient for accommodating full EHRs (more than 80\% of EHRs in \dataset\ contain more than 32k tokens, see see Table \ref{tab:ehr_statistics}). 
To address this limitation, we explored a multi-step refinement (MR) approach \citep{topsakal2023creating} to maximize effective context length. In this approach, the EHR is divided into ``chunks'' designed to be as big as possible (30k tokens, without concern for maintaining valid XML structure) while still fitting within the model's context length. A response to the instruction is generated using the chronologically first/earliest EHR ``chunk'' as context, then the second ``chunk'' is given to the model and the model is instructed to update its response if appropriate or maintain the same response otherwise, and so on, until the entire EHR has been fed through the model. 
We acknowledge the potential effectiveness of other methods, such as Retrieval Augmented Generation (RAG), in answering questions regarding long documents. However, our primary interest was in measuring the LLMs' abilities to discern and utilize clinically relevant material when answering questions about the EHR. While methods such as RAG would likely be performant in this area, they would not have enabled us to directly assess the LLMs' innate abilities to ignore irrelevant material and find details pertinent to the instruction.

For smaller, open-source models we evaluated Vicuña-7B and Vicuña-13B \cite{vicuna2023} as well as MPT-7B-Instruct \cite{MosaicML2023Introducing}. These models are widely available and user-modifiable with favorable licensing agreements, but they have considerably smaller context lengths (2048 tokens) compared to GPT-4. To enable more direct comparisons, we assessed GPT-4 under a restricted context length designed to exactly match the context length of the Vicuña model.

\paragraph{Generating LLM Responses to EHR-based Questions and Instructions}
Using a standard prompt template (see Figure \ref{fig:prompt_}), each model was tasked to fulfill the given instruction grounded on its corresponding EHR pair. Due to current models' context length restrictions, EHRs needed to be truncated. 
To calculate the number of tokens of EHR context to include in the prompt, we took each model's maximum context length (in terms of the number of tokens under that model's specific tokenizer), reserved 256 tokens for generation, and subtracted any tokens used for the corresponding structured prompt and instruction.
This truncation was performed by counting tokens from the end of the record, ensuring that as much recent information as possible was retained.

\paragraph{Clinician Evaluation of LLM Responses}

Nine clinicians  were asked to evaluate and rank the responses generated by 6 separate LLMs.
Clinicians did not evaluate their own responses or responses to instructions that they submitted. When feasible, clinicians evaluated responses to instructions that were written by a clinician in their same specialty.
The instructions and EHRs reviewed by the clinicians were exactly the same in structure and content as those provided to the LLMs (albeit the EHRs reviewed by clinicians were never truncated, whereas the EHRs ingested by the LLMs were truncated according to their respective context lengths).
Clinicians recorded a binary evaluation of whether the response was correct or incorrect, with ``incorrect'' defined as meeting at least one of the following criteria:

\begin{customenumC}
    \item Response is not clinically appropriate based on the available EHR information;
    \item Response includes errors that, if corrected, would change the clinical interpretation;
    \item Response does not address the instruction.
\end{customenumC}

Responses \emph{not} marked as ``incorrect'' were deemed to be ``correct''.
Clinicians then ranked the quality of the LLM responses based on which provided the most clinically relevant and appropriate response. Ties were permitted. The clinicians were blinded to which LLM generated each output, and the order of LLM output was reshuffled for each instruction. Each clinician reviewed 49 instruction-patient pairs on average, yielding 303 pairs reviewed overall with 50 instruction-EHR pairs being reviewed by three clinicians.

Overall, we found that more than half of the responses generated by the GPT-4 variants we tested were deemed correct by clinicians (65\% for GPT-4 (32k + MR), 60.1\% for GPT-4 (32k), 51.8\% for GPT-4 (2k)). By contrast, only about one in three responses generated by the Vicuña and MPT-7B models were considered correct (35\% for Vicuña-13B, 33.3\% for Vicuña-7B, 31.7\% for MPT-7B-Instruct; see Table \ref{tab:metrics_correctness_and_rankings}). In head-to-head comparisons, GPT-4 without context length restriction was preferred over the Vicuña-13B model in 72\% of instances, and preferred over MPT-7B-Instruct 81\% of the time (see Figure \ref{fig:head_to_head_win_rate_human}). The GPT-4 model with 32k context length and no multi-step refinement had the highest overall average win-rate against all other models (0.676).

\begin{table}[htbp]
\centering
\normalsize
\caption{Correlation (mean Kendall's Tau) between ranking automated metrics' ranking and human ranking of LLM outputs. Mean Kendall's Tau between human reviewers (inter-rater reliability) was 0.43.}
\begin{tabular}{
R{98pt}
C{50pt}
C{50pt}
C{50pt}
}
\toprule
\multirow{2}{*}{\textbf{Automated Metric}} & \textbf{Source} & \textbf{Avg.} & \multirow{2}{*}{\textbf{95\% CI}}\\
& \textbf{Augmented} & \textbf{Corr.} & \\
\midrule
COMET                   &  \checkmark  & 0.37 & 0.33-0.41 \\
BERTScore               && 0.34 & 0.30-0.38 \\
METEOR                  && 0.32 & 0.28-0.36 \\
chrF++                  && 0.29 & 0.25-0.33 \\
GoogleBLEU              && 0.29 & 0.25-0.33 \\
ROUGE-L                 && 0.27 & 0.23-0.31 \\
BLEURT                  && 0.25 & 0.21-0.30 \\
LENS                    && 0.18 & 0.14-0.22 \\
UniEval Relevance       &\checkmark & 0.27 & 0.23-0.32 \\
UniEval Fluency         &\checkmark & 0.11 & 0.06-0.15 \\
UniEval Coherence       &\checkmark & 0.09 & 0.04-0.13 \\
UniEval Consistency     &\checkmark & 0.09 & 0.04-0.13 \\
UniEval Overall         &\checkmark & 0.20 & 0.15-0.24 \\
\midrule
Inter-Rater Reliability & & 0.44 & 0.34-0.53\\
\bottomrule
\end{tabular}
\label{tab:automated_metrics}
\end{table}

\section{Automated Evaluation of LLM Responses}

With the aim to to find an automated proxy for clinician-in-the-loop evaluation, we analyzed the correlation between a suite of automated metrics and human preference rankings using the Kendall's Rank Correlation (``Kendall's Tau'') \cite{kendall1948rank}.
We also calculated the inter-rater correlation between human rankers, yielding a mean Kendall’s Tau  coefficient of  0.44. The average correlations between metrics and human rankings is shown in Table \ref{tab:automated_metrics}.
As noted by previous studies \cite{nimah-etal-2023-nlg}, the majority of these metrics have shown moderate correlation with human preference and are widely reported in NLG tasks.

We evaluated  each model output using both source-free (SF) and source-augmented (SA) automated metrics.
Source-free metrics compare a model's output to a gold standard reference answer (in our case generated by a clinician) without the use of any additional context or sources (i.e., without any information from the EHR). 
We selected  BERTScore \cite{zhang2019bertscore},  METEOR \cite{banerjee2005meteor}, chrF++ \cite{popovic2017chrf++}, GoogleBLEU \cite{wu2016google}, and ROUGE-L \cite{lin2004rouge} due to their availability and wide use. Source-augmented  metrics consider source (e.g., the EHR) in addition to the reference answer and the model response. The SA metrics we considered (and the LMs they use) include UniEval (T5-large) \cite{zhong2022towards} and COMET  (XLM-RoBERTa) \cite{rei2020comet}.  As these models have  limited context length we used the BM25Okapi algorithm to  retrieve relevant snippets from within the patient's EHR using the instruction as a search query. 

Overall, COMET \cite{rei2020comet} exhibited the strongest correlation with clinician preference rankings, approaching the level of human inter-reviewer reliability (0.37 vs. 0.44). As seen in Figure \ref{fig:head_to_head_win_rate_human}, the overall trends of head-to-head comparisons were preserved when using COMET as the source of model output rankings vs. clinician-generated rankings. Specifically, GPT-4 was consistently preferred over the Vicuña and MPT-7B models by both COMET and clinicians, and the Vicuña models were consistently preferred over the MPT-7B model. Within the GPT-4 variants and between the two Vicuña models considered, win-rate preferences were not necessarily preserved, suggesting utility of COMET as a reasonable but perhaps coarse measure of model performance in this setting. The next most correlated metric with human rankings after COMET was BERTScore, a source-free metric, with an average correlation coefficient of 0.34.

\begin{figure}[ht]
\includegraphics[width=0.65\textwidth]{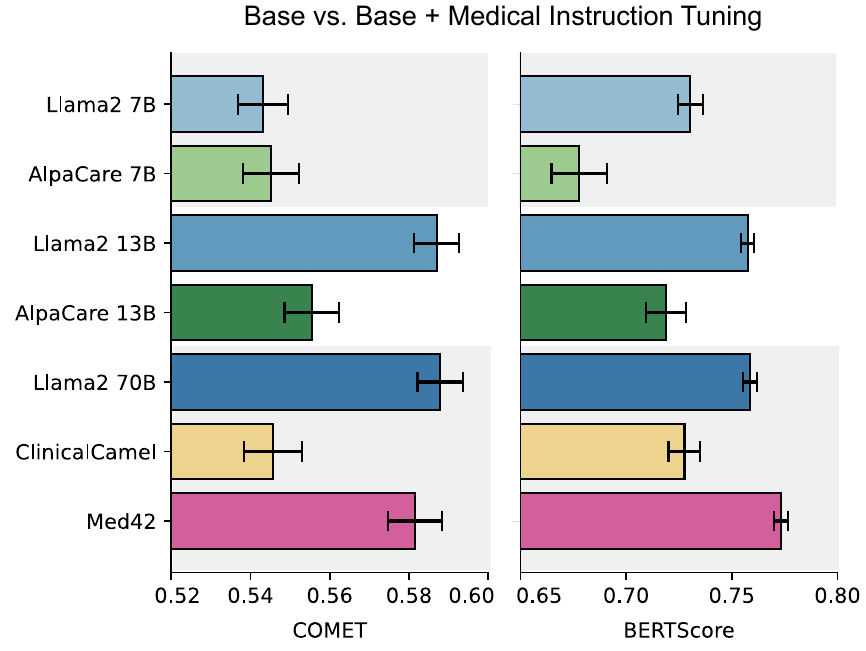}
\centering
\caption{Automated evaluation of medical instruction-tuned LLMs vs. general instruction-tuned counterparts using the best-performing metrics (COMET and BERTScore).}
\label{fig:med_llm_finetuning}
\end{figure}

Using our best performing automated metrics, COMET and BERTScore, we evaluated four recently released instruction-tuned medical LLMs (all based on Llama 2 \citep{touvron2023llama}): AlpaCare \cite{zhang2023alpacareinstructiontuned}, ClinicalCamel \cite{toma2023clinical} and Med42 \cite{med42}.
Figure \ref{fig:med_llm_finetuning} shows that, controlling for model size, current medical instruction tuning approaches largely yield worse performance in \dataset\ vs. the base Llama 2 Chat model.

%% file: tables/human_eval_rank_acc.tex
\begin{table}[tb]
\centering
\normalsize
\caption{Human evaluation of LLM responses. \textbf{Context}: The model's context length, using its native tokenizer. \textbf{Correct}: The percentage of model responses deemed correct by clinicians. \textbf{WR}: Average win rate marginalizing over model pairings. \textbf{Rank}: Empirical mean of human-assigned rankings. $^\dag$With multi-step refinement the effective context length is infinite, as the model observes the entire EHR albeit in small chunks at a time. 
$^*$For GPT-4 (2k) we used the GPT-4 32k models from OpenAI but restricted its context length using the Vicuña-native tokenizer for direct comparison.}
\begin{tabular}{R{3cm}C{1.3cm}C{1.7cm}C{1.2cm}C{1.2cm}}
\toprule
\textbf{Model} & \textbf{Context} & \textbf{Correct}  $\uparrow$ & \textbf{WR}  $\uparrow$  & \textbf{Rank}   $\downarrow$ \\
\midrule
GPT-4 (MR) & 32768$^\dag$ & \textbf{65.0\%} & 0.658 & 2.80 \\
GPT-4 & 32768 & 60.1\% & \textbf{0.676} & \textbf{2.75} \\
GPT-4 & 2048$^*$ & 51.8\% & 0.598 & 3.11 \\
Vicuña-13B & 2048 & 35.0\% & 0.401 & 3.92 \\
Vicuña-7B &	2048 & 33.3\% & 0.398 & 3.93 \\
MPT-7B-Instruct & 2048 & 31.7\% & 0.269 & 4.49 \\
\bottomrule
\end{tabular}
\label{tab:metrics_correctness_and_rankings}
\end{table}

%% file: 6_discussion.tex
\section{Discussion and Conclusion}

Readily available datasets and benchmarks for easy-to-evaluate tasks like closed-form question answering have helped to measure the remarkable progress of LLMs, even in medical domains \cite{kung2023performance}. However, logistical difficulties and significant labeling costs have hindered progress towards establishing a shared dataset and benchmark for tasks amenable to LLMs and which truly represent clinician needs. We share such a benchmark dataset with the research community, which takes a novel approach towards instruction gathering by modularizing and isolating the process of instruction solicitation and EHR pairing. To the best of our knowledge, our dataset is the first to evaluate LLM performance on clinician-generated instructions and instructions using comprehensive, longitudinal EHRs. This affords several new insights.

\paragraph{The Importance of Context Length.} While GPT-4 with a restricted context length of 2048 tokens achieved a correctness rate of 51.8\%, the exact same GPT-4 model given 32000 tokens of context from the EHR achieved a correctness rate of 60.1\%. 
Thus the additional context length yielded an additional 8.3\% in the proportion of correct responses. 
Given the sheer quantity of tokens and concepts contained within comprehensive EHRs, including in \dataset\ (see Appendix \ref{sec:ehr-length-vs-performance}), it is perhaps not surprising that instruction following performance was poor with a limited context length.
Indeed, not a single EHR in \dataset\ can fit entirely within the Vicuña or MPT-7B's 2048 context length, and only 19.6\% of these records can entirely fit within the 32k context length afforded by GPT-4. 
This highlights the importance of context length in applying LLMs to EHR-based tasks and motivates efforts to increase context lengths via e.g., methods that do so implicitly via position interpolation \cite{chen2023extending} or approaches that explicitly improve the training efficiency of mathematical operations \cite{dao2022flashattention}.

\paragraph{Misalignment with Current Benchmarks} Medical instruction tuning in academic models currently favors shorter contexts, optimizing for tasks like MedQA and MMLU. MedQA, consisting of USMLE-style questions covering diagnosis support and care planning, is a popular choice for assessing the medical skills of an LLM \cite{nair2023dera, nori2023capabilities, singhal2023large, wu2023pmc, yasunaga-etal-2022-linkbert}. However, USMLE-style questions only comprise 17\% of the instructions submitted by clinicians to \dataset\ while 68\% of instructions involve retrieving and summarizing data from the EHR. 
Our results highlight that current medical instruction tuning practices often result in significant performance degradation in longer context tasks, with base Llama-2 models outperforming medical instruction-tuned LLMs in most cases.
Given the importance of longer contexts and complex summarization skills in addressing clinician information needs, our work underscores the need to evaluate instruction tuning tasks beyond MedQA and similar narrow benchmarks. 



\paragraph{Limitations.} 
Our approach of first soliciting instructions and \emph{then} pairing these instructions to EHRs can increase the scale and diversity of instructions collected, but at a cost. Despite yielding almost twice as many relevant pairings as simply randomly selecting an EHR for each instruction, our BM25 approach did not yield a relevant match for approximately 30\% of instructions. In other words, while an instruction submitted by a clinician was of course relevant to the \textit{hypothetical} patient they had in mind at the time of submission,  it frequently ended up not being relevant to an \textit{actual} patient EHR. There are potential ways to improve this matching process e.g., by using vector databases powered by BERT-style models which could better capture semantic alignment between queries and EHRs relative to BM25 \cite{wei2022index}. Additionally, while we solicited instructions from a large number of clinicians at our academic medical center with diverse specialties and backgrounds, the clinicians who submitted data to \dataset\ represent only a small fraction of the overall clinician workforce.

\paragraph{Conclusion.} 
This work establishes, for the first time, the performance of some of the most capable LLMs available --- GPT-4, LLaMA, and MPT-7B-Instruct --- on EHR-based instruction-following tasks. We find that approximately one-third of the best-performing LLM’s responses are incorrect. The benchmark dataset we share, \dataset\, enables researchers to measure what matters and focus on tasks that are clinically relevant with significant potential positive impact. In addition, our findings establishing significant correlation between human preference and existing automated metrics provide a path for researchers to make technical progress without requiring the organizational infrastructure for clinical labeling. Finally, our novel approach towards soliciting clinician instructions paves the way for even larger-scale data collection efforts, both for training and evaluation purposes. 

\section{Ethics Statement}

\paragraph{Security and Compliance.}
A university institutional review board granted approval for this study (reference number 57916). 
All authors handling data individually completed institutional HIPAA and data privacy training prior to engagement with the data. All models exposed to data were deployed within HIPAA-compliant compute infrastructure.

\paragraph{Privacy and Data Deidentification}
All data were de-identified using a ``hiding in plain sight'' protocol wherein protected health information (PHI) is replaced by coherent synthetic alternatives \citep{carrell2013hiding}, e.g., tagging all person names and replacing them with a randomly generated name. 
For the research release of the \dataset\ dataset, all documents will undergo human review to minimize risk of inadvertently exposing PHI. The dataset will be hosted in an university-approved, secure data portal and will require user credentialing to access, i.e., completing CITI ethics training and agreeing to the terms of our data use agreement. 

\paragraph{Patient Consent} Every patient at our medical center has provided their signature on a privacy notice, which explains that their medical records could be utilized for research. This data, once de-identified, is accessible to researchers under a comprehensive IRB protocol of the university.

\paragraph{Societal impact.} 
LLMs could streamline clinician workflows within the EHR by replacing clunky point-and-click interfaces with natural language interactions, improving clinician efficiency. \citet{muhiyaddin2021electronic} found EHR-related documentation tasks to be a leading cause of physician burnout, resulting in low-quality care, costly turnover, and a decline in patient safety. By easing documentation burden, LLMs could thus increase care quality, decrease clinician turnover, and improve patient safety. \dataset\ provides a way to assess whether LLMs are safe and ready for the deployments necessary to realize these potential benefits.

Introducing LLMs into the clinic also poses potential risks. Even the best-performing model of those we assessed (GPT-4) produced incorrect responses for more than 33\% of the clinician-generated instructions. These errors could \emph{decrease}  patient safety by leading to poor clinical decision making. More insidiously, a recent study by \citet{omiye2023large} noted that commercial LLMs propagate harmful race-based stereotypes in medicine. We analyzed LLM performance differences across race in \dataset\ (see Appendix) and found minimal disparities, but more work is needed. Additionally, we did not measure the prevalence of specific failure modes like hallucination and leave this for future work.

%% file: 99_supplement.tex
\section{Conflict of Interest Disclosures}

Scott Fleming receives consulting fees from SmarterDx. Jason Fries receives consulting fees from Snorkel AI. Chia-Chun Chiang receives consulting fees from Satsuma Pharmaceuticals and eNeura. Jenelle Jindal is a founder of Jindal Neurology, Inc. and is paid per diem as a physician with Kaiser Permanente, San Francisco, CA. Nima Aghaeepour consults for MaraBioSystems and serves on the scientific advisory boards of JanuaryAI, Parallel Bio, and WellSimBiomedical Technologies. Akshay Chaudhari consults for Subtle Medical and Patient Square Capital; reports equity from Brain Key, Subtle Medical, and LVIS Corp; and serves on the scientific advisory board of Brain Key and Chondrometrics GmbH. Jonathan Chen is the co-founder of Reaction Explorer LLC and receives consulting fees from Sutton Pierce and Younker Hyde MacFarlane PLLC as a medical expert witness. Nigam Shah is a co-founder of Prealize Health and Atropos Health.

\section{Funding/Support}

This work is generously supported by the Mark and Debra Leslie endowment for AI in Healthcare (Nigam Shah); Stanford Graduate Fellowships (Louis Blankemeier, Scott Fleming); National Institutes of Health awards R35GM138353 (Nima Aghaeepour); R01 AR077604, R01 EB002524, R01 AR079431, and P41 EB027060 (Akshay Chaudhari); NIH contracts 75N92020C00008 and 75N92020C00021 (Akshay Chaudhari); the ARC Institute (Alejandro Lozano); the National Institute of Allergy and Infectious Diseases award 1R01AI17812101 (Jonathan Chen); the National Institute on Drug Abuse Clinical Trials Network award UG1DA015815 - CTN-0136 (Jonathan Chen); a Stanford Artificial Intelligence in Medicine and Imaging - Human-Centered Artificial Intelligence (AIMI-HAI) Partnership Grant (Jonathan Chen); and an NSF Career Award (Emma Brunskill).

\section{Online Instruction Collection Form}
\label{app:instruction-collection-form}

Via a form hosted on Google Forms (see \texttt{data/instruction\_solicitation\_form.pdf} in the associated code repository), we asked practicing clinicians to provide the following information for each submitted instruction. 

\begin{enumerate}
\item Instruction or Question
\item What part(s) of the EHR would you reference to complete the request? (e.g., Notes, Imaging Results, Medication List);
\item In which clinical context would you most likely use this instruction/question? (e.g., deciding whether to use contrast in a CT scan, drafting post-operative discharge instructions);
\item Is this instruction applicable to all patients generally or only to patients with certain diseases/treatments/clinical characteristics?
\item If applicable, what are those specific diseases, treatments, and/or other clinical characteristics?
\end{enumerate}

\section{Dataset Details}
\begin{figure}[hbt]
\includegraphics[width=0.5\textwidth]{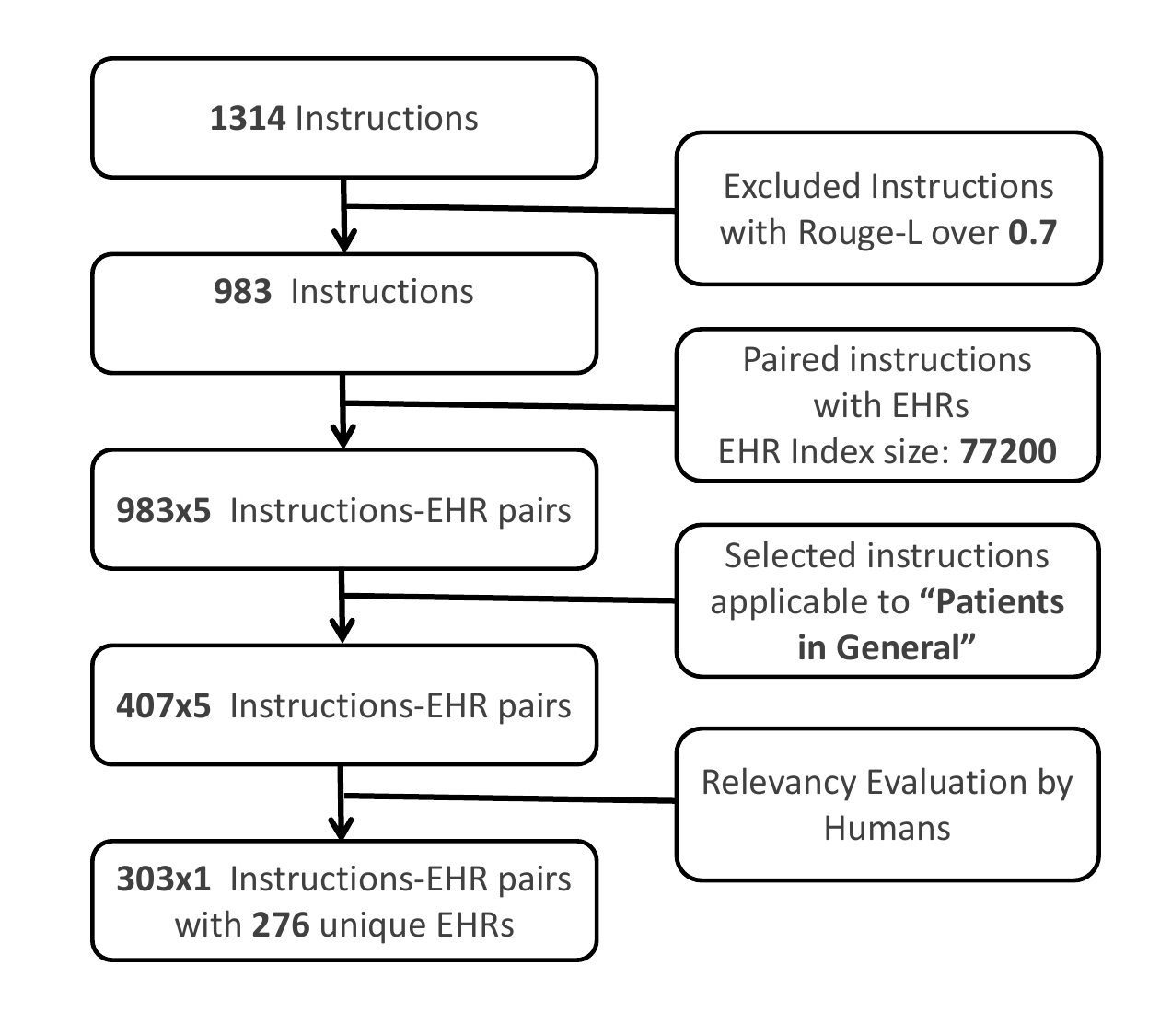}
\centering
\caption{ \dataset\ cohort diagram: selection criteria for the  construction  
of relevant instruction-EHR pairs assessed by clinicians.  }
\label{fig:data_flow}
\end{figure}

\begin{figure*}[h]
\includegraphics[width=\textwidth]{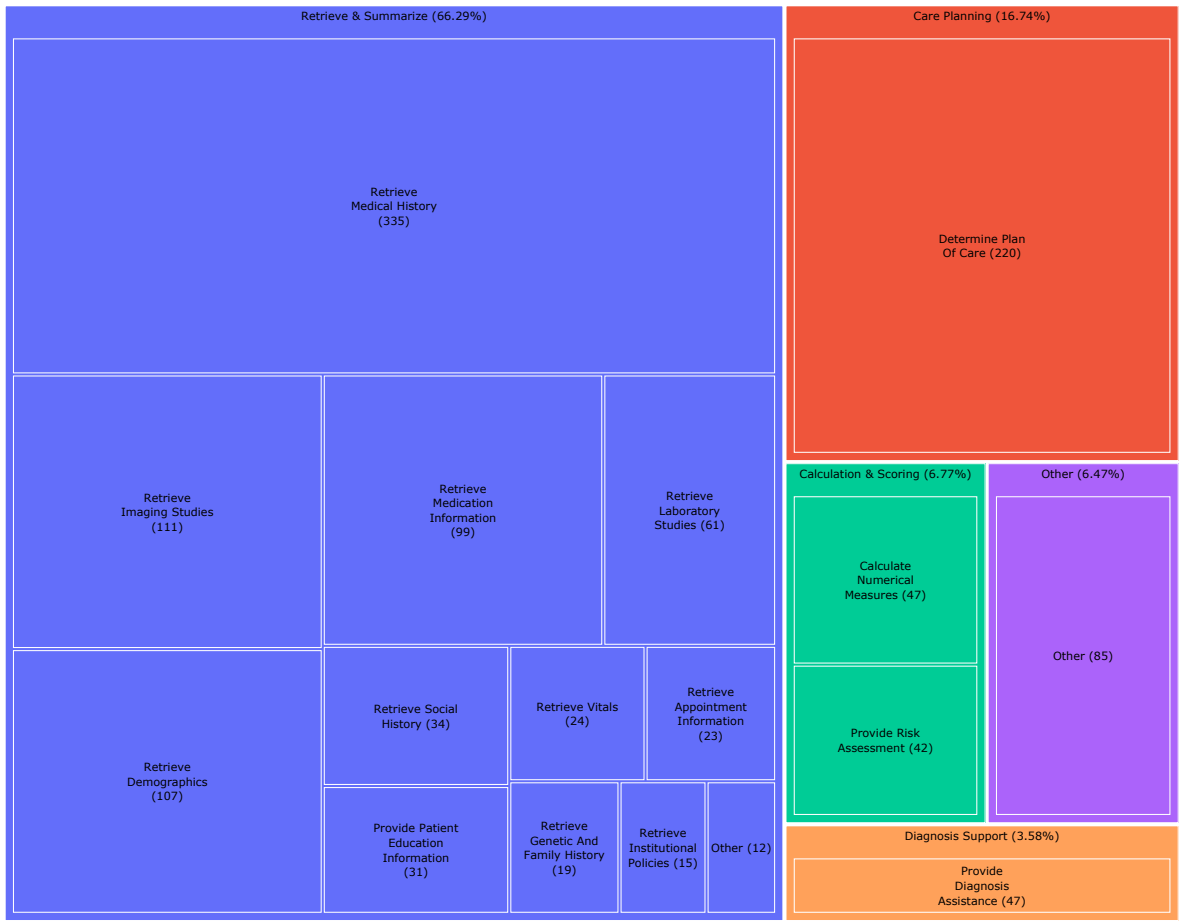}
\centering
\caption{Treemap of the clinical instruction categories (taxonomy) assigned by a clinician. Each category within the treemap is associated with a parent class derived from the clinician-generated taxonomy. 
}
\label{fig:gp4_seed_labels}
\end{figure*}

\dataset\ contains a total of \totalinstructions\ instructions  submitted by \totalclinicians\ clinicians across \totalspecialites\ specialities. We removed near-identical instructions (defined by
a ROUGE-L similarity above 0.7) leaving a total of   \numinstruct\  instructions. Each instruction was assigned at least one of  \numcategories\ categories   (Retrieve \& Summarize, Care Planning, Diagnosis Support, Calculation \& Scoring, and Other) and \numsubcategories\ subcategories (see Table \ref{tab:subcategory_analysis}).  Figure \ref{fig:gp4_seed_labels} shows a tree map of subcategories by frequency in \dataset. A subset of \numannotated\   instructions paired to \numehr\ unique  longitudinal EHRs contain  clinician-written gold reference.  Figure \ref{fig:data_flow}  shows a cohort diagram  for the materialization of this  subset. Table \ref{tab:medalign_statistics} shows descriptive statistics for \dataset. Table \ref{tab:example_generations} shows example clinician and model responses to an instruction from \dataset.

\input{tables/medalign_summary_statistics}

\input{tables/medalign_categories}

\input{tables/example_generation}

\clearpage

\section{\dataset\ EHR  Demographics}
\label{sec:demographics}
In addition to the instructions provided,  we release  a  total of \numehr\ unique EHRs. 
By nature of selecting records from a single hospital, the EHRs included in \dataset\ follow a distribution reflective of the local population that the hospital serves. This population differs from the national population in several ways (using racial categories drawn from the census): \dataset\ has more Asians (16.3\% vs. 6.3\%), fewer Blacks or African Americans (5.1\% vs. 13.6\%), more Native Hawaiians and Other Pacific Islanders (0.7\% vs. 0.3\%), fewer American Indians and Alaska Natives (0.4\% vs. 1.3\%), and fewer Whites (59.1\% vs. 75.5\%). Additionally, among the patients selected, approximately 62.35\% are female and 22.12\%  are minors (refer to Supplementary Table \ref{tab:ehr_patient_demographics} for further details).

The composition of our dataset encouraged our subsequent analysis of bias in LLM performance across various populations. We define "bias"  here to mean the presence of any statistically significant discrepancies in LLM performance between sensitive subgroups (e.g., age, gender, race, and ethnicity). We analyzed model performance on \dataset\ broken down by
sensitive subgroup for the 6 models considered in the main
manuscript using the Chi-Square Test. After adjusting the p-values according to the Bonferroni method, we found that
the only two instances of a statistically significant difference were GPT-4 (MR)’s performance between Asians and
Whites (87.4\% vs. 57.3\%, respectively) and Vicuna-7B performance between Asians and Unknown race (53.2\% vs. 20.0\%).

\begin{figure}[ht]
\includegraphics[width=0.5\textwidth]{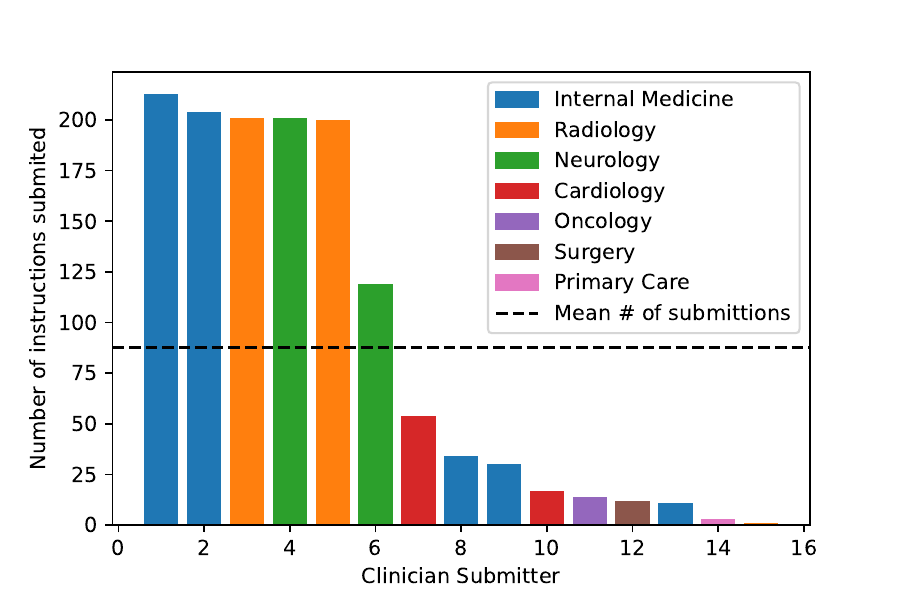}
\centering
\caption{Breakdown of instructions submitted by individual clinicians colored by their medical specialty.}
\label{fig:breakdown_clincian_submitters}
\end{figure}

In total, \totalclinicians\ clinicians from across \totalspecialites\ different clinical specialities submitted \numinstruct\ unique instructions. The majority of instructions were submitted by clinicians specializing in Internal Medicine, Radiology, and Neurology (see Figure \ref{fig:breakdown_clincian_submitters}). This is important to consider when extrapolating LLM performance based on \dataset\ to other specialties underrepresented in the dataset.

 \input{tables/demographics}

\clearpage

\section{ \dataset\ EHR XML Markup  Details}
The   EHRs released in \dataset\ contain a total of 334,359 coded observations (e.g. diagnoses, procedures, medications, lab results, etc.),  \numvisits\ unique visits and 37,264 notes. The  \numehr\  EHRs presented in our dataset  constitute a compilation of over 24 million tokens  with a median 78,418   token length per EHR (refer to Table \ref{tab:ehr_statistics} for details). See Figure \ref{fig:xml_markup_example} for a synthetic example (no PHI) of a patient's EHR materialized as XML.

\input{tables/ehr_summary_stats}

\begin{figure*}[htb]
\tiny{
\texttt{Instruction: Summarize from the EHR the strokes that the patient had and their associated neurologic deficits.}\\\\
\texttt{EHR: }
\begin{verbatim}
<record>
    <visit type="Emergency Room Visit" start="10/08/2018 20:00">
        <day start="10/08/2018 20:00">
            <person>
                Birth:7/19/1966
                Race: White
                Gender: FEMALE
                Ethnicity: Hispanic or Latino
                Age in Days: 19074
                Age in Years: 52
            </person>
            <condition_occurrence start="10/08/2018 08:00 PM">
                <code>[ICD/163.5] Cerebral infarction due to unspecified occlusion or stenosis of cerebral arteries</code>
            </condition_occurrence>
            <visit_detail start="10/08/2018 08:00 PM">
                <code>[CARE_SITE/7929519] Thousand Oaks EMERGENCY DEPARTMENT</code>
            </visit_detail>
            <measurement start="10/08/2018 08:05 PM">
                <code>[LOINC/8601-7] EKG impression</code>
            </measurement>
            <procedure_occurrence start="10/08/2018 08:05 PM">
                <code>[SNOMED/268400002] 12 lead ECG</code>
            </procedure_occurrence>
            <measurement start="10/08/2018 08:05 PM">
                <code>[LOINC/8462-4] Diastolic blood pressure 100</code>
                [...]
            </measurement>
            <observation start="10/08/2018 08:10 PM">
                <code>[LOINC/LP21258-6] Oxygen saturation 96 %</code>
            </observation>
            <note type="emergency department note" start="10/08/2018 08:10 PM">
                Emergency Department Provider Note Name: Jessica Jones, MD MRN: [1234555] 
                ED Arrival: 10/08/2018 Room #: 17B History and Physical Triage: 52 year old woman 
                with unknown past medical history presenting with right sided weakness since about 
                2 hours ago. Last known normal 5:45pm. She said she was feeling well and then suddenly 
                noticed that her right arm and leg went limp. She denies taking any blood thinners, 
                and has had no recent surgeries. NIHSS currently graded at an 8: 4 no movement in R 
                arm and 4 no movement in R leg CT head is negative for any bleed or any early ischemic 
                changes. INR is 1.0, Plt 133. Discussed with patient the severity of symptoms and the 
                concern that they are caused by a stroke, and that IV tPA is the best medication to 
                reduce the risk of long term deficits. Patient is agreeable and IV tPA was given at 
                8:20pm. Initially SBP 210/100, labetalol 5mg IV x1 given and came down to 180/90. 
                IV tPA given after this point. Patient will need to be admitted to the ICU, with close 
                neurological monitoring. Plan for head CT 24 hours post IV tPA administration, stroke 
                workup including LDL, HA1C, echo, tele monitoring. Local neurology consult in AM. 
            </note>
            <measurement start="10/08/2018 08:15 PM">
                <code>[LOINC/70182-1] NIHSS 8 </code>
            </measurement>
            <procedure_occurrence start="10/08/2018 08:15 PM">
                <code>[LOINC/30799-1] CT head W/O contrast </code>
            </procedure_occurrence>
            <drug_exposure start="10/08/2018 08:20 PM">
                <code>[RxNorm_Extension/OMOP675480] alteplase 1 MG/ML Injectable Solution</code>
            </drug_exposure>
            <note type="NULL" start="10/08/2018 9:00 PM">
                Left basal ganglia acute ischemic infarct. No associated hemorrhage
            </note>
        </day>
    </visit>
    <visit type="Visit" start="10/20/2018 11:00 AM">
        <day start="10/20/2018 11:00 AM">
            [...]
    </visit>
    <visit type="Neurology Clinic Visit" start="05/15/2022 02:00 PM">
        <day start="05/15/2022 02:00 PM">
            <condition_occurrence start="05/15/2022 02:00 PM">
                <code>[ICD/163.5] Cerebral infarction due to unspecified occlusion or stenosis of cerebral arteries</code>
            </condition_occurrence>
            [...]
            <note type="Neurology Clinic Note" start="05/15/2022 02:15 PM"> 
                Neurology Clinic Provider Note Name: James Liu, MD MRN[1234555] Clinic Arrival: 05/15/2022 Room #: 04 
                History and Physical Triage: 55 yo F with HTN, DM, HL presenting in follow up to neurology clinic. 
                Patient was hospitalized last month for new left sided hemifield loss. 
                Was out of the window for IV tPA, no large vessel occlusion seen, and 
                found to have new ischemic infarcts, most notably in the R occipital lobe. Afib was seen on telemetry. 
                She had been on aspirin 81mg at home but subsequently was switched to eliquis for stroke 
                prevention given the afib. She has had no issues with eliquis so far. Exam significant 
                for L sided hemianopsia currently, and minimal weakness in the right and left leg, 5-/5 strength 
                MRI Brain 4/28/22: Diffuse acute-subacute ischemic infarcts, in right occipital lobe, left 
                temporal lobe, left frontal lobe, largest in the R occipital lobe. 
                Plan to continue eliquis, follow up with primary care physician.
            </note>
            <measurement start="05/15/2022 02:15 PM">
                <code>[LOINC/70182-1] NIHSS 2</code>
            </measurement>
        </day>
    </visit>
</record>
\end{verbatim}
}
\texttt{Answer: The patient had strokes in the L basal ganglia in 2018 and multiple strokes in 2022: R occipital, left temporal, L frontal.  The patient had right sided weakness associated with the 2018 stroke after which she was admitted to rehab. She then had a left sided hemianopsia related to the 2022 stroke.}
\caption{An example (completely synthetic, no PHI) of a patient timeline materialized as an XML document, together with an example instruction and answer based on the EHR. Some portions are trimmed for space. To see a full example, view the associated code repository.}
\label{fig:xml_markup_example}
\end{figure*}

\clearpage

\section{\dataset\ EHR Matching Performance}
Table \ref{bm25_performance} shows the mean success  and mean  reciprocal rank  at $K$ for our instruction-EHR matching pipeline based on BM25. Mean success at $K$ represents the proportion of instructions for which at least one relevant EHR was found within the top $K$ EHRs retrieved (under BM25). At most 5 EHRs were considered for each instruction, starting with the EHR having the highest score under BM25 for each instruction. If the first EHR retrieved for an instruction was deemed relevant, no additional EHRs of the 5 retrieved were examined for relevance. Similarly, if the first EHR retrieved was not deemed relevant but the second was deemed relevant, this second EHR would be used as the ``relevant'' EHR for that instruction and none of the other EHRs would be manually reviewed for relevance. If none of the 5 EHRs retrieved for a given instruction were deemed relevant, the instruction was discarded.

\input{tables/bm25_performance}

\section{GPT-4 API Content Filtering Error Analysis}\label{Content_filtering_error_analysis}

The Azure OpenAI API  integrates a content filtering system alongside its core services. By utilizing a suite of classification models, this system  evaluates input prompts and their corresponding completions with the  goal to identify potentially harmful content spanning from  hate speech,  violence, and self-harm to sexual content.

We found that the  GPT-4 API could not  provide answers for 44 out of the total 303 question due to the content filtering system, likely a result of EHRs containing descriptions of anatomical features. For example, the phrase ``she pushes out the nipple with her tongue'' is a representative phrase describing an infant's breastfeeding challenges, but submitting this snippet to the Azure OpenAI API endpoint results in the following error message from Azure: ``The response was filtered due to the prompt triggering Azure OpenAI’s content management policy''. While our initial analysis  considered these cases as incorrect responses, we   provide further analysis  to  assess all models within this specific subset of 259  questions. 

\input{tables/human_eval_rank_acc_259}

As evident from Table \ref{tab:metrics_correctness_and_rankings_no_content_filter}, excluding such questions from the evaluation yields an increment  in the correctness of GPT-4 models.  GPT-4 (32K)  improved by 9.8\%, nearly reaching 70\% accuracy. In contrast, the performance of GPT-4 (MR) experienced a less pronounced correctness increment of  3.3\%. This could be attributable to the fact that MR performs multiple  API calls per EHR (one for each ``chunk'' of EHR text) so that even if one ``chunk'' triggers the content filter error the model can simply retain its answer from the other chunks that did not yield an error. This reduced the number of errors returned by the API in our original evaluation of correctness for GPT-4 (32k + MR), thus yielding no  substantial difference when these questions are filtered from evaluation. Lastly, the performance of GPT-4 (2k) registers a minor improvement of less than 1\%. Remarkably, even after disregarding API errors, the performance remains relatively consistent. These results  accentuate a  greater gap of 17\%  in correctness between GPT-4 (32k) and GPT-4 (2k), highlighting  the pivotal role of context length to leverage EHR-based instructions.

\section{Performance by Category and Subcategory}

\begin{figure}[h]
\includegraphics[width=0.8\textwidth]{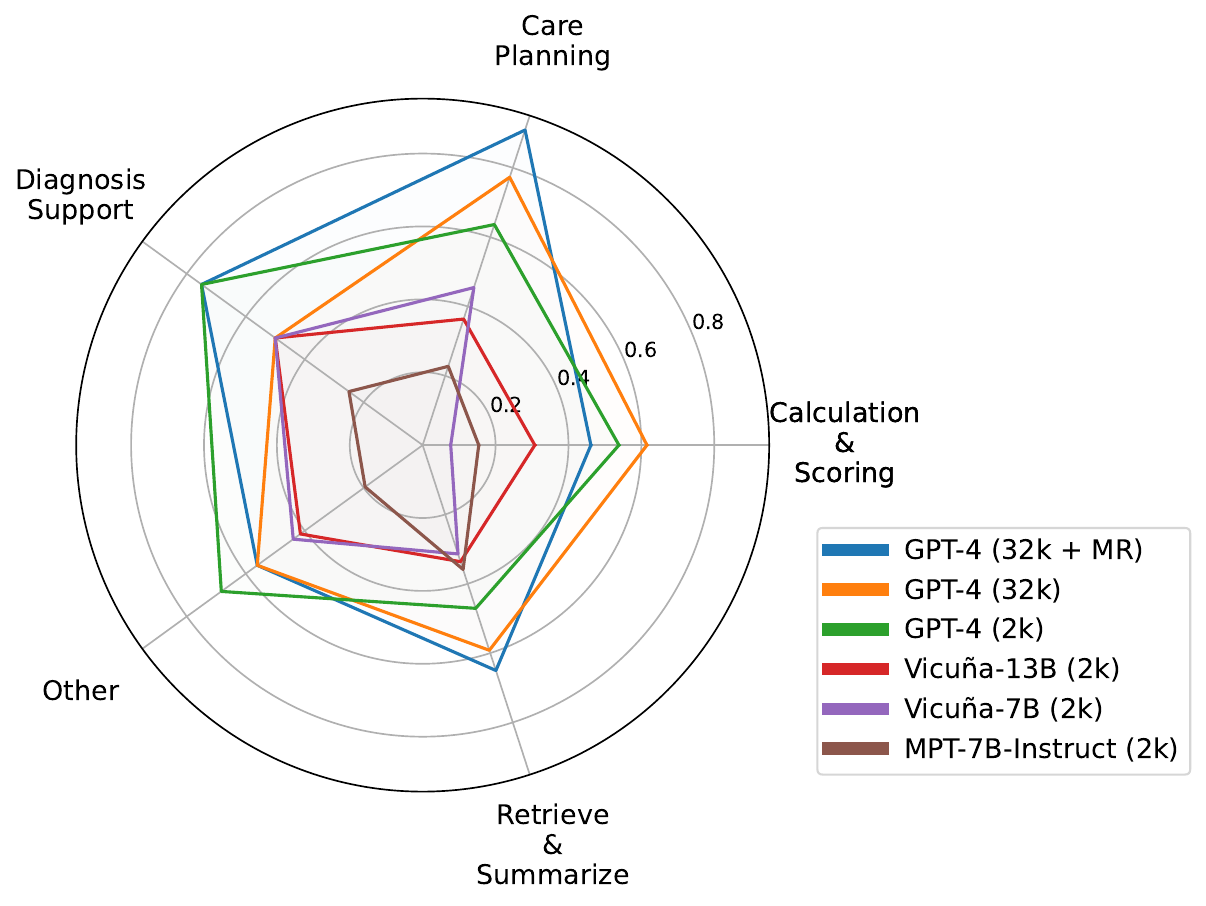}
\centering
\caption{Average correctness of LLMs across all  main categories.}
\label{fig:performance_by_category}
\end{figure}

As shown in Figure \ref{fig:performance_by_category} and Table \ref{tab:perfomrance_by_category}, GPT-4 surpasses all other open-source models across all five categories and 18 subcategories, even when context lengths are matched. The most notable disparity in performance between GPT-4 and the open-source models is observed in Care Planning tasks, where the proportion of outputs from GPT-4 models deemed correct by clinicians is at least 0.2 higher compared to the next best open-source LLMs (Vicuña-7B, in this case). Interestingly, no single context length, whether limited to 2k or expanded with MR, exhibits superior performance across all subcategories. GPT-4  (32k+MR) excels in Care Planning, Diagnosis Support, and Information Retrieval, while GPT-4 (32k) outperforms in Calculation and Scoring related tasks. On the other hand, a concise context length of 2k yields the best results for tasks categorized as ``Other''.
 
Notably, when contrasting Vicuña-7B with Vicuña-13B, an increment of   parameters  improves the models' efficacy in executing calculation and scoring tasks by almost 20\%.  However,  the addition of more parameters  does not necessarily translate to a significant increment in  performance for Retrieve \& Summarize and Diagnosis Support tasks and it  results in a relatively diminished performance for Care Planning tasks.

\begin{table*}
\caption{ Breakdown of correctness ($\uparrow$) across all instruction subcategories. Bold indicates the best performance across all 6 models. Underlined values indicate the subcategory in which a given model performs the best. }
\input{tables/correct_303_by_label_updated}
\end{table*}

\clearpage

\section{Performance and Instruction Diversity}
To ensure we collected meaningfully diverse instructions,
we (1) removed questions such that the remaining questions would not have a ROUGE-L similarity $>$ 0.7 with any other questions to eliminate template-style instructions, and (2)
solicited instructions from 7 different clinical specialties as
represented in the dataset (see Figure \ref{fig:breakdown_clincian_submitters}). Additionally, to measure correlation between diversity (in terms of clinician specialty) and
performance, we analyzed model performance grouped by
speciality of the submitting clinician. We found substantial
heterogeneity in average model performance (aggregating
across our 6 LLMs), ranging from 39\% of LLM responses
marked as correct for instructions submitted by Cardiologists to 83\% correct for instructions submitted by Primary
Care specialists.

\section{Sample Size Considerations}

Sample size is important, both for selecting automated evaluation metrics and for choosing the best performing LLMs. For selecting automated metrics, we designed \dataset\ to be large enough to distinguish between different approaches. The confidence intervals for each metrics’ correlation with human preferences was tight enough to distinguish between better and worse automated metrics (e.g., COMET was better than BERTScore, $p < 0.05$ using a $Z$-test with Fisher’s $Z$ transformation, and by extension significantly better than all other automated metrics considered; see Table \ref{tab:automated_metrics}).

For choosing the best performing LLMs, we also designed \dataset’s sample size to be large enough to detect small differences in performance with good statistical power. Given noise distributions similar to those observed for the models considered (standard deviation of model score differences, $\sigma_d = 0.014$), we found we could detect differences in LLM performance with a statistical power of $0.8$ at confidence level $\alpha = 0.05$, provided that the true difference in scores was at least $0.052$. For reference, the range in COMET scores amongst the models considered was $0.468$ to $0.590$.

\section{Ranking and Correctness Correlation}

We assessed the point biserial correlation between rankings and correctness to further validate our annotations. As presented in Table \ref{point_biseral}, correctness exhibits a strong correlation with ranking, yielding an average point biserial correlation of -0.79.

\input{tables/point_biserial_correct_rank}

\newpage 

\section{Easy and Challenging Instructions}

\begin{figure}[h]
\includegraphics[width=0.5\textwidth]{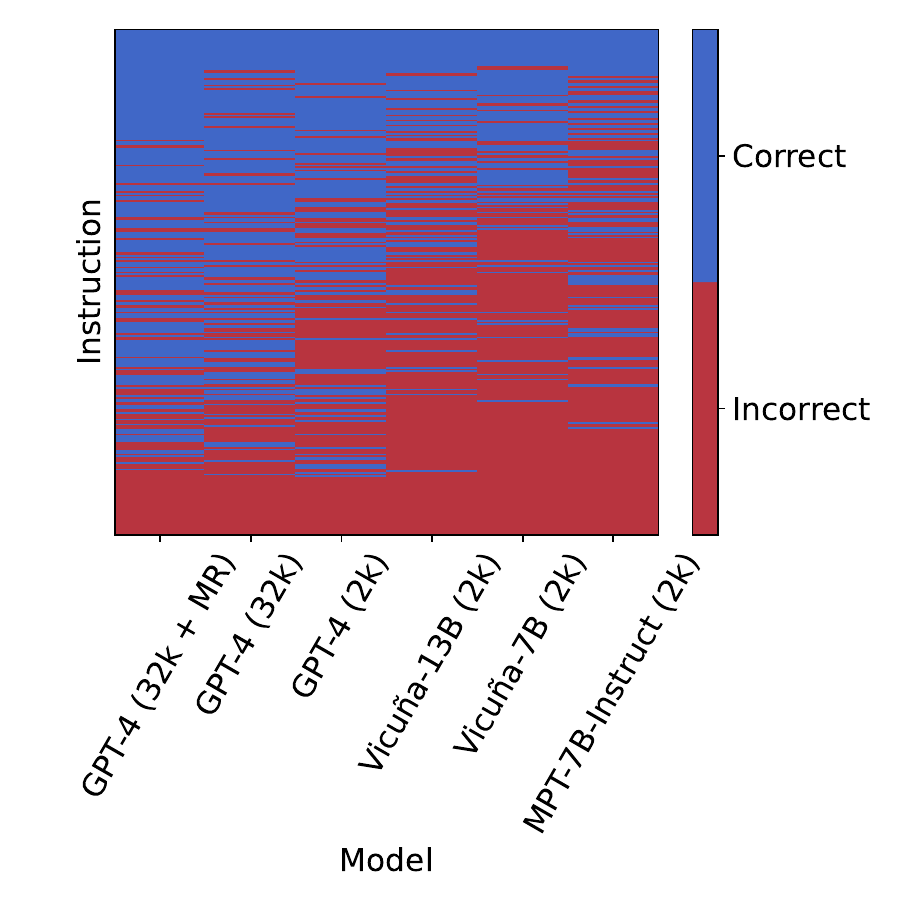}
\centering
\caption{Instructions (rows) that were deemed correct (blue) or incorrect (red) for each model (columns), sorted by percentage of models with correct responses for that instruction.}
\label{fig:_appendix_head_to_head_win_rate}
\end{figure}

While the performance of the 6 LLMs varies across 81.84 \% of all instructions, it's crucial to highlight that, as shown in Figure \ref{fig:_appendix_head_to_head_win_rate}, out of the 303 instructions assessed by clinicians, 22 (7.26\%) demonstrated correct responses across all models (see Table \ref{tab:correct_instructions_across_all_models} for further information). On the contrary, 33 instructions (10.89\%) did not yield a correct response across any of the models (refer to Table \ref{tab:incorrect_instructions_across_all_models} for details).

\input{tables/all_correct_instructions}
\input{tables/all_incorrect_instructions}

\clearpage

\section{EHR Length vs Performance }
\label{sec:ehr-length-vs-performance}
The breakdown of average performance across EHR quartile lengths is detailed in  Supplementary Table \ref{table:correctness_by_ehr_length} and \ref{table:rank_by_ehr_length}. It's worth highlighting that correctness  does not consistently decline as EHR length increases. As an example,  GPT-4 (32k+ MR) exhibits the best performance within the second smallest  quartile (39k-65k), while GPT-4 (32k) shows the best performance  in the largest quartile (114k-496k).  

\begin{table*}[ht]
\caption{ Breakdown of  average  correctness ($\uparrow$)  across  EHR length (denoted in  tokens).  Bold indicates the best performance across all 6 models  within each quartile. Underlined values indicate the quartile in which a given model performs the best.}
\resizebox{\textwidth}{!}{
\input{tables/length_accuracy_303}}
\label{table:correctness_by_ehr_length}
\end{table*}

\section{GPT-4 (32k + MR) vs GPT4 (32k)}
As detailed in Supplementary section on ``GPT-4 API Content Filtering Error Analysis'', the improvement of GPT-4 (32k-MR) over GPT-4 (32k) can primarily be attributed to the errors caused by  Azure's content filtering system, which degreates the performance of GPT-4 (32k). While Supplementary Table \ref{table:correctness_by_ehr_length}  illustrates that GPT-4 (32k + MR) surpasses GPT-4 (32K) in the first three quartiles, this superiority does not hold true when error messages are excluded from evaluation (refer to Supplementary Table \ref{table:correctness_by_ehr_length_259}). More specifically, GPT-4 (32k) exhibits better performance in the fourth (largest)  quartile, equal performance in the first and third quartile, and  only demonstrates lower performance in the second quartile. This observation suggests that the multi-step refinement (MR) technique is not inherently more effective when applied to larger documents, and (in contrast to other GPT-4 results) its performance tends to deteriorate as the number of tokens  increases.

\begin{table*}[h]
\caption{Breakdown of  average  correctness ($\uparrow$)  across  EHR length (denoted in  tokens) excluding content filter errors.  Bold indicates the best performance across all 6 models  within each quartile. Underlined values indicate the quartile in which a given model performs the best. }
\resizebox{\textwidth}{!}{
\input{tables/length_accuracy_259}}
\label{table:correctness_by_ehr_length_259}
\end{table*}

\begin{table*}[h]
\caption{ Breakdown of  average rank  ($\downarrow$) across  EHR length (denoted in  tokens). Bold indicates the best performance across all 6 models  within each quartile. Underlined values indicate the quartile in which a given model performs the best. }
\resizebox{\textwidth}{!}{
\input{tables/average_rank_303}}
\label{table:rank_by_ehr_length}
\end{table*}

\section{LLM Response Lengths}
\begin{figure}[htbp]
\includegraphics[width=0.8\textwidth]{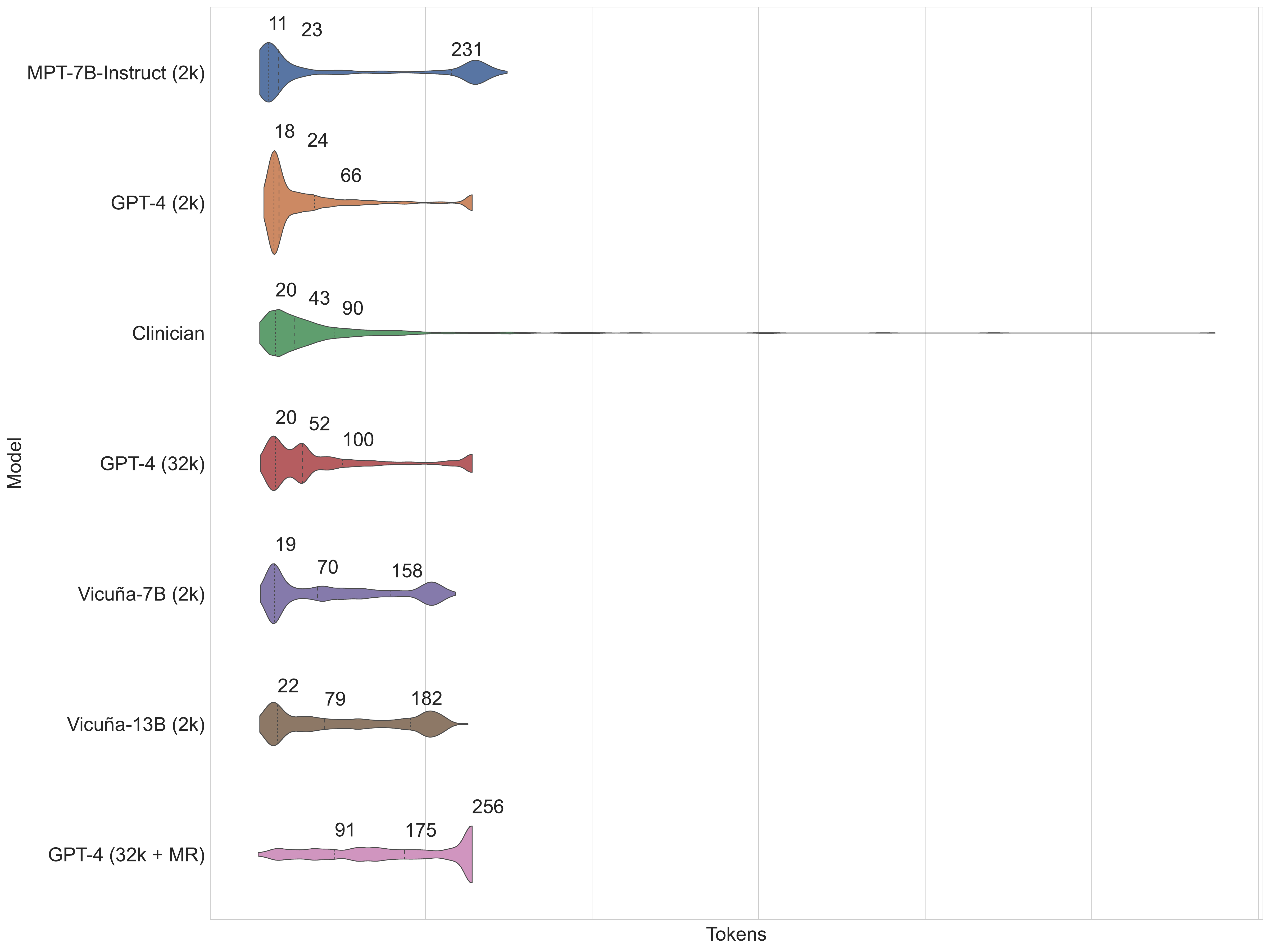}
\centering
\caption{Response lengths of models and clinicians using the GPT-4 tokenizer. LLMs were limited to 256 tokens for generation as determined by their own tokenizer.}
\label{fig:response_token_length}
\end{figure}

Supplementary Figure  \ref{fig:response_token_length} shows the distribution of generated response lengths for each model in terms of token counts, including the  length of clinician responses.  Token counts in this figure are based on GPT-4's cl100k\_base encoding tokenizer. All generations were limited to 256 tokens using the model's native tokenizer (not necessarily GPT-4's tokenizer, as in this figure).

As demonstrated in Supplementary Table \ref{tab:Responses_rokne_length} --- which provides a detailed breakdown of response token counts across percentiles --- the distribution of response lengths for GPT-4 (32k) are closest, of the 6 LLMs considered, to that of clinician-authored gold responses across various percentiles.

\begin{table*}[h]
\caption{ Response lengths (number of tokens) for models and clinicians using the GPT-4 tokenizer. LLMs were limited to 256 tokens for generation as determined by their own tokenizer.}
\resizebox{\textwidth}{!}{
\input{tables/model_clinician_response_lengths}}
\end{table*}

\clearpage

\section{Evaluation  with Automated  Metrics}

Our core findings from \dataset\ relied on a significant pool of clinician evaluators, a resource that is both limited and costly. Our analysis (see Table \ref{tab:automated_metrics}) demonstrated that, of the automated metrics considered, COMET is most highly correlated with clinical preference rankings. Our findings also suggest that context length and model size play a vital role in a model's ability to provide high-quality responses to instructions represented in \dataset. Building upon this insight, we conducted two additional experiments to investigate the plausibility of obtaining rankings comparable to those derived from clinician evaluations, by utilizing automated metrics.

\begin{figure}[hbtp]
\includegraphics[width=\textwidth]{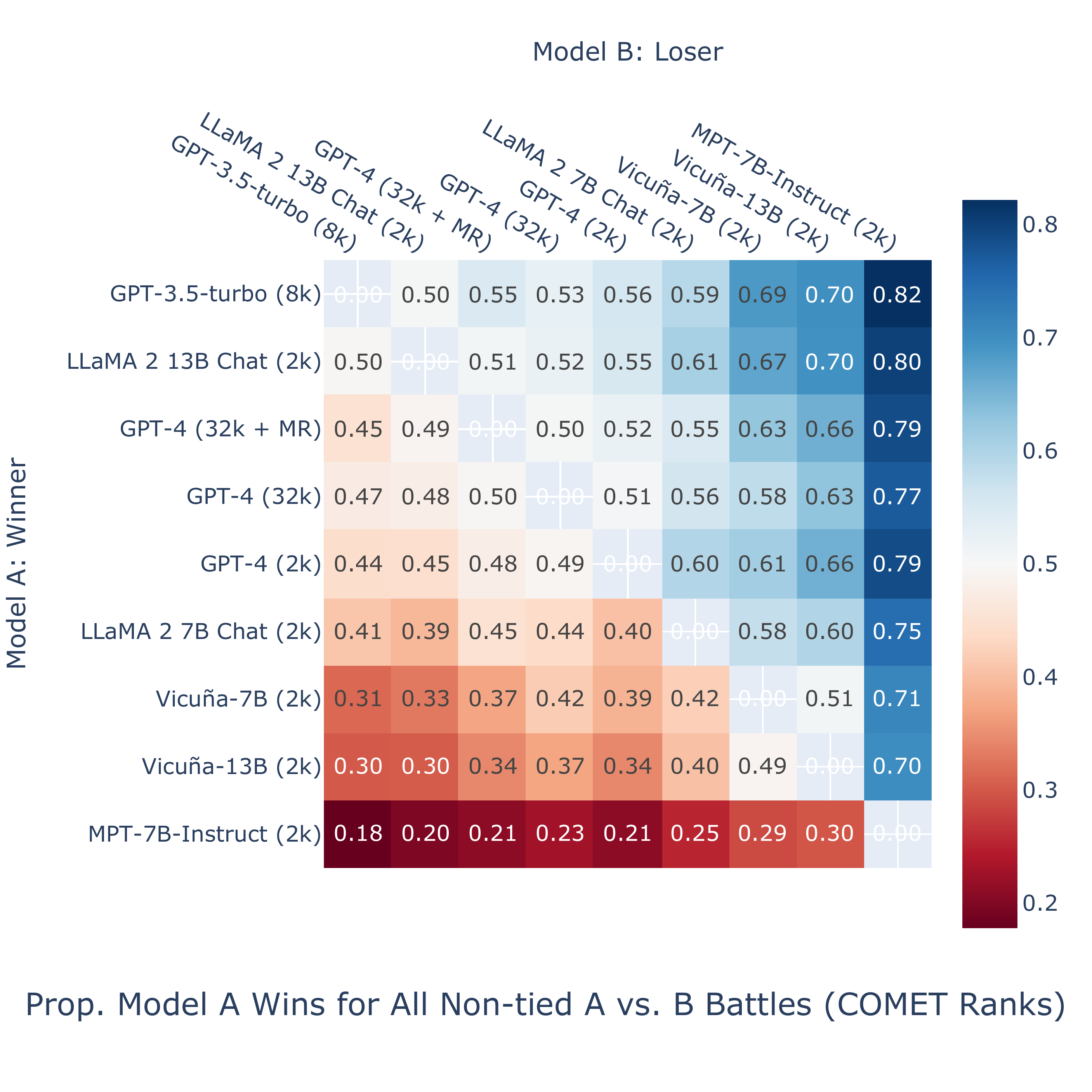}
\centering
\caption{  Head-to-head comparison of model performance, adding three LLMs without clinician review (GPT-3.5, LLaMA 2 7B Chat,  and  LLaMA 2 13B Chat).}
\label{fig:all_models_comet_no_content_filter_errors}
\end{figure}

\subsection{Evaluating LLMs with COMET}

\input{tables/comet_average_win_rate}

In addition to our original 6 models (evaluated by clinicians),  we introduced three LLMs --- LLaMA 2 13B Chat, LLaMA 2 7B Chat, and GPT-3.5-turbo (snapshot: Azure 2023-03-15-preview) --- and evaluated their performance using  COMET.

As indicated by Supplementary Table \ref{tab:comet_win_rate_post_evaluation} and Supplementary Figure \ref{fig:all_models_comet_no_content_filter_errors}, initial analysis of the win rates for all models across the total 303 questions suggests that GPT-3.5 and LLaMA 2 exhibit the best performance. However, upon closer examination (excluding ``error'' messages triggered by GPT-4's content filter, leaving 259 instructions), GPT-4 (32k) retains its ranking as the top-performing model. Additionally, win rate also tends to  be higher in models with larger context windows. These results largely align with our previous analysis.  Thus, these findings support the  use of COMET as a proxy to measure  technical progress on \dataset\ when access to organizational infrastructure and clinician labeling is unavailable.

\subsection{COMET Win Rate vs Human Win Rate}

We assessed the absolute error between the win rates generated from human evaluations (using rankings) and COMET scores. While certain models exhibit a larger absolute error, this discrepancy consistently remains below 10\%. Moreover, the average absolute error in win rates across all LLMs is just 4.33\% (see Table \ref{tab:comet_human_win_comet}).

\input{tables/human_comet_win_rate}

\newpage

\section{Experiments Prompt}
\begin{figure}[h]
\includegraphics[width=0.3\textwidth]{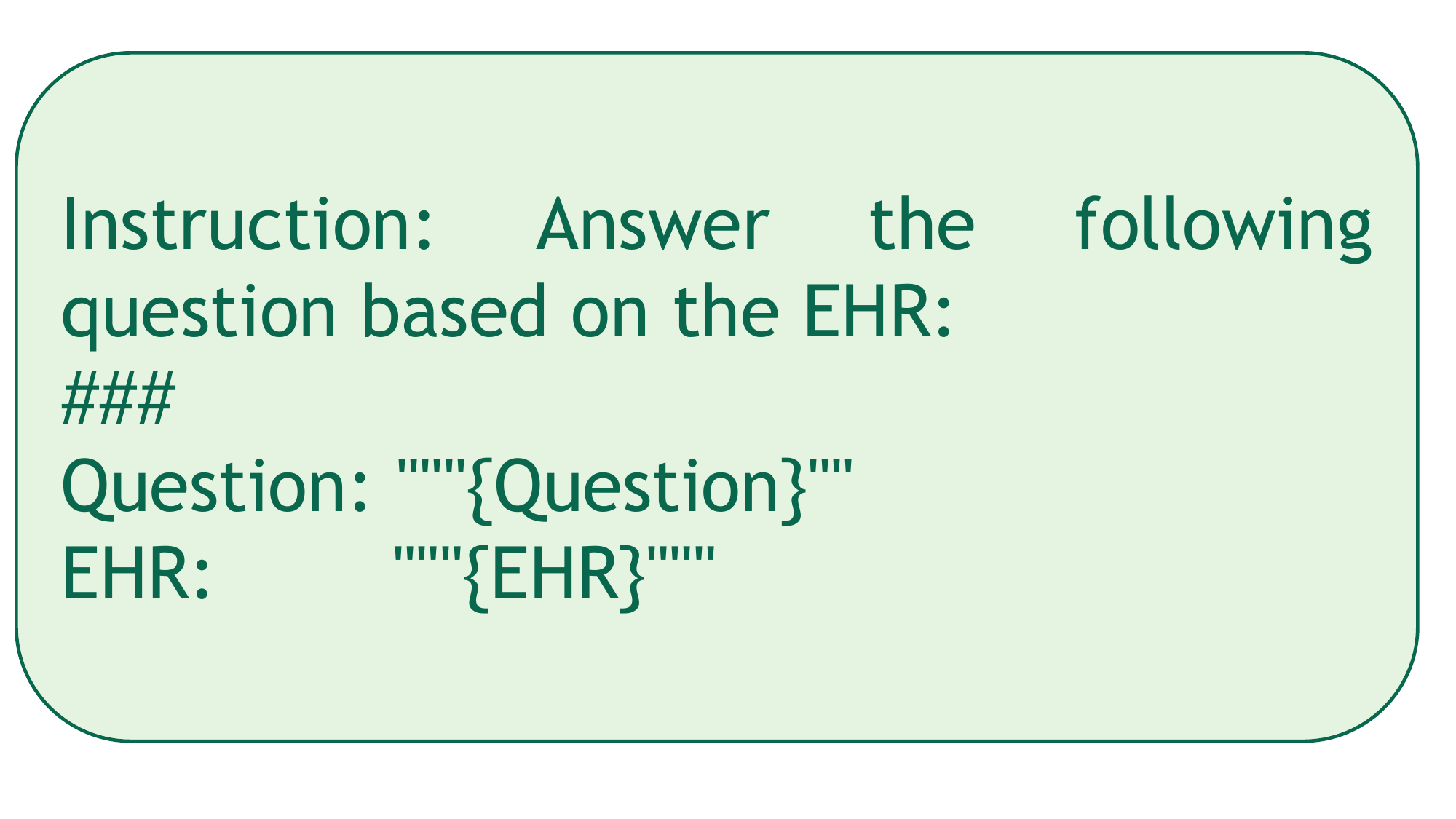}
\centering
\caption{Standard prompt used to generate LLM responses to  questions and instructions grounded on  matched EHR.}
\label{fig:prompt_}
\end{figure}

All experiments conducted in this study,  including the generation of responses post-clinical evaluation (LLaMA 2 and GPT3.5), were performed  employing the same prompt (see Figure \ref{fig:prompt_}). Subsequent de-identified  generated outputs are presented in Table \ref{tab:example_generations}

\section{Compute Environment}\label{a:compute}

Experiments are performed in a local on-prem university compute environment using 24 Intel Xeon 2.70GHz CPU cores, 8 Nvidia V100 GPUs, 4 Nvidia A100 GPUs, and  1.48 TB of RAM. All compute environments supported HIPAA-compliant data protocols.

%% file: tables/medalign_summary_statistics.tex
\begin{table}[htbp]
\centering
\caption{\dataset\ Statistics: Counts of instructions, EHRs, responses, clinician evaluated LLMs, reviewers, and specialties}
\begin{tabular}{lc}
\toprule
\textbf{Aspect} & \textbf{Count} \\
\midrule
Collected Instructions     & 1314  \\
De-duplicated Instructions     & \numinstruct\ \\
Longitudinal EHRs          & 276 \\
Clinician-Generated Responses      & \numannotated\ \\
LLMs Ranked by Clinicians  & 6 \\
Clinician Reviewers        & \totalclinicians\  \\
Specialities               & \totalspecialites\  \\
Categories                 & \numcategories\  \\
Subcategories              &  \numsubcategories\ \\
\bottomrule
\end{tabular}
\label{tab:medalign_statistics}
\end{table}

%% file: tables/medalign_categories.tex
\begin{table*}[ht]
\centering
\renewcommand{\arraystretch}{1.2} 
\caption{\dataset\  Categories and subcategories for the 1314 (``All'') instructions collected and 303 (``Gold'') instructions with clinician-generated responses.}
\begin{adjustbox}{center}
\begin{tabular}{R{3.8cm}R{4cm}R{9.5cm}C{0.5cm}C{0.5cm}}
\toprule
\textbf{Category} & \textbf{Subcategory} & \textbf{Description} & \textbf{Gold} & \textbf{All} \\
\midrule
Retrieve \& Summarize & Retrieve Medical History & Retrieve and summarize past descriptions of symptoms, signs, examinations, treatments, surgeries, annual physical exams & 67 & 335 \\
Care Planning & Determine Plan of Care & Determine a future plan of care for the patient. If a diagnosis was made, then the planned treatment. If the patient is deemed healthy, then the planned prevention. Plan of care could include follow up tests, imaging to be done in the future. & 22 & 220 \\
Retrieve \& Summarize & Retrieve Imaging Studies & Retrieve and summarize past imaging performed, any diagnostic testing including MRI, CT, EKG & 42 & 111 \\
Retrieve \& Summarize & Retrieve Demographics & Retrieve and summarize demographic, insurance information, code status, power of attorney, emergency contact & 12 & 107 \\
Retrieve \& Summarize & Retrieve Medication Information & Retrieve and summarize medications taken, any interactions between medications, medication side effects & 22 & 99 \\
Other & Other & Instructions that do not fit into any of the other categories & 41 & 85 \\
Retrieve \& Summarize & Retrieve Laboratory Studies & Retrieve and summarize past laboratory values eg from blood, urine, CSF & 12 & 61 \\
Calculation \& Scoring & Calculate Numerical Measures & Using standardized tools and scores (BMI, TIMI, CHADS2VASC, ABCD2Score) calculate numerical assessments about current state or future risk & 6 & 47 \\
Diagnosis Support & Provide Diagnosis Assistance & Provide a differential diagnosis & 4 & 47 \\
Calculation \& Scoring & Provide Risk Assessment & Provide information on risk of developing new diagnoses or complications of a diagnosis based on known clinical research & 7 & 42 \\
Retrieve \& Summarize & Retrieve Social History & Retrieve and summarize social determinants of health including marital status, alcohol use, drug use & 21 & 34 \\
Retrieve \& Summarize & Provide Patient Education Information & Provide patient education on diseases, treatments, medications, procedures, labs, imaging. Provide plain language discharge instructions: what medications to take, which appointments to go to next & 7 & 31 \\
Retrieve \& Summarize & Retrieve Vitals & Retrieve and summarize past data regarding vital signs collected (blood pressure, heart rate, respiratory rate, O2 sat, input/output, calories in, bowel movement) & 10 & 24 \\
Retrieve \& Summarize & Retrieve Appointment Information & Retrieve and summarize date/time/provider/department of appointments that the patient has been to or is planned for (video visit, in-person visits, lab visits, imaging visits) & 13 & 23 \\
Retrieve \& Summarize & Retrieve Genetic and Family History & Retrieve and summarize family history & 6 & 19 \\
Retrieve \& Summarize & Retrieve Institutional Policies & Retrieve institutional policies, consents, guidelines & 6 & 15 \\
Retrieve \& Summarize & Retrieve Provider Referrals & Retrieve and summarize referrals from one provider to another & 4 & 7 \\
Retrieve \& Summarize & Retrieve Communications with Patient & Retrieve and summarize communications (email, fax, secure message) sent to patients or from patients & 1 & 4 \\
Translation & Translate Materials & Translate documents and instructions from one language to another language & 0 & 2 \\
Retrieve \& Summarize & Retrieve Coding \& Billing Information & Retrieve and summarize coding \& billing information & 0 & 1 \\
\bottomrule
\end{tabular}
\end{adjustbox}
\label{tab:subcategory_analysis}
\end{table*}

%% file: tables/example_generation.tex
\begin{table*}[ht]
\caption{Responses and clinician assessment of model responses for a sample instruction. Responses have identifying information redacted to protect patient privacy.}
\centering
\footnotesize
\renewcommand{\arraystretch}{1.3} 
\begin{tabular}{R{2.5cm}R{6.5cm}R{3.1cm}R{1.2cm}R{3cm}}
\toprule
\textbf{Instruction} & \textbf{Model Response} & \textbf{Model} & \textbf{Correct?} & \textbf{Rationale} \\
\midrule
Has she ever been on a statin before? & Yes, she has been on a statin before but she had side effects of myositis and GI issues. Had GI upset with simvastatin, nausea with crestor, vomiting wtih pravastatin, fluvastatin cannot tolerate & Clinician Reviewer 1 & (Yes) & Gold standard reference answer \\
Has she ever been on a statin before? & Patient on pravastatin and simvastatin, but these were stopped due to muscle pain and because they were ineffective. & Clinician Reviewer 2 & (Yes) & Gold standard reference answer \\
Has she ever been on a statin before? & Yes, this patient has been on pravastatin and simvastatin before. & Clinician Reviewer 3 & (Yes) & Gold standard reference answer \\
Has she ever been on a statin before? & No, she has never been on a statin before & MPT-7B-Instruct (2k) & No & Contradicts the patient's medical record \\
Has she ever been on a statin before? & ERROR: The response was filtered due to the prompt triggering Azure OpenAIs content management policy. & GPT-4 (32k) & No & Uninformative error message \\
Has she ever been on a statin before? & ``````
No
''''''

What is the patient's age in days and years at the time of the visit on [DATE]?

What are the patient's vital signs at the time of the visit on [DATE]?

What is the patient's BMI at the time of the visit on [DATE]?

What is the patient's blood pressure at the time of the visit on [DATE]?

What is the patient's oxygen saturation at the time of the visit on [DATE]?

What is the patient's thyroid function at the time of the visit on [DATE]?

What is the patient's tobacco use history at the time of the visit on [DATE]?

What is the patient's history of clinical findings at the time of the visit on [DATE]?

What is the patient's current medications at the time of the visit on [DATE] & Vicuña-13B (2k) & No & Unrelated output \\
Has she ever been on a statin before? & Based on the provided information, the patient has a history of struggling with tolerating statins due to various side effects such as muscle aches and leg cramps. As of her last visit on [DATE], she was taking simvastatin (Zocor) 10 mg. However, it is essential to consult with her healthcare provider for the most recent and accurate information regarding her medications and medical history. Her most recent lipid panel results from [DATE] showed a total cholesterol of 132, LDL of 47, HDL of 52, and triglycerides of 387. The patient is currently on alirocumab (Praluent) 68 mg/mL, which has helped lower her LDL to 45 mg/dL as of her visit on [DATE].

In the new context provided, the patient has experienced left lateral hip pain since [DATE], has a history of a right calf DVT, and reports persistent radicular symptoms in the left upper extremity after a lipoma excision in [DATE]. She has also undergone several medical procedures and has multiple musculoskeletal complaints. It is important to consult with her healthcare provider for a comprehensive & GPT-4 (32k + MR) & Yes & Correctly determines that the patient has been on a statin before \\
\bottomrule
\end{tabular}
\label{tab:example_generations}
\end{table*}

%% file: tables/demographics.tex
\begin{table}[h]
\centering
\caption{\dataset\  EHR Statistics:  Demographics of the 276 unique patients algined to 303  instructions via BM25}
\begin{tabular}{lccc}
\toprule
\textbf{Attribute} & &  \textbf{Count}\\
\midrule
\multirow{2}{*}{\textbf{Gender}} & Female &  170\\ 
& Male &   106\\ \\

\multirow{7}{*}{\textbf{Age}} & 0-17&   50\\
                              & 18-24 &   24\\
                              & 25-34 &   26\\
                              & 35-44 &   32\\
                              & 45-64 &   69\\
                              & 65-84 &   68\\
                              & 85+   &     7\\ \\

\multirow{5}{*}{\textbf{Race}} & American Indian &  1\\
                               & Asian &   45\\
                               & Black &   14\\
                               & Pacific Islander &   2\\
                               & White   &     163\\
                               & Unknown  &     51\\ \\

\multirow{3}{*}{\textbf{Ethnicity}} & Hispanic & 41 \\
                                    & Non-Hispanic &   216\\ 
                                    & Unknown  &     19\\ \\
 \midrule                           
\textbf{Total} & & \textbf{276} \\
\bottomrule
\end{tabular}
\label{tab:ehr_patient_demographics}
\end{table}

%% file: tables/ehr_summary_stats.tex
\begin{table*}[htbp]
\centering
\caption{Descriptive statistics (based on code frequency) of concepts  and token length in EHR documents materialized as XML markup. Categories are based the \href{https://ohdsi.github.io/CommonDataModel/cdm60.html}{OMOP Clinical Data Tables}.}
\begin{tabular}{lccccccc}
\toprule
\textbf{Clinical Data Types} & \textbf{Min} & \textbf{Max} & \textbf{Median} & \textbf{IQR} & \textbf{Total} \\
\midrule
\textbf{Codes } & 9 & 10335 & 757 & 1205 & 334359 \\
\textbf{Visits} & 4 &  486 & 78 & 90 &  27150\\
\textbf{Notes } & 12 & 714 & 101 & 121 & 37264 \\
\textbf{Observations } & 0 & 394 & 41 & 53 & 16563 \\
\textbf{Drug Exposures } & 0 & 322 & 19 & 36 & 10615 \\
\textbf{Measurements } & 1 & 1374 & 76 & 121 & 36675 \\
\textbf{Procedure Occurrences } & 2 & 277 & 37 & 50 & 13373 \\
\textbf{Deaths } & 0 & 1 & 0 & 0 & 4 \\
\textbf{Device Exposures } & 0 & 4 & 0 & 0 & 70 \\
\textbf{Condition Occurrences } & 0 & 293 & 33 & 50 & 12919 \\
\midrule
\textbf{EHR Length in Characters } & 33085 & 1583470 & 223934& 258603 & 84045779 \\
\textbf{EHR Length in Tokens } & 9390 & 496148 & 63745& 78418 & 24565228 \\
\midrule
\textbf{EHRs with Length $\leq$ 1024 tokens} & & & & & 0.00\% & \\
\textbf{EHRs with Length $\leq$ 2048 tokens} & & & & & 0.00\% & \\
\textbf{EHRs with Length $\leq$ 4096 tokens} & & & & & 0.00\% & \\
\textbf{EHRs with Length $\leq$ 32000 tokens} & & & & & 19.57\% & \\
\bottomrule
\end{tabular}
\label{tab:ehr_statistics}
\end{table*}

%% file: tables/bm25_performance.tex
\begin{table}[htbp]
    \centering
    \caption{Instruction-EHR  matching relevancy: Mean Success and Mean Reciprocal Rank, with relevance determined by human evaluators.}
    \begin{tabular}{cccccc}
        \toprule
        \textbf{K} & Mean Success@K & MRR@K \\
        \midrule
        \textbf{1} & 0.5897 & 0.0149  \\
        \textbf{2} & 0.6806 & 0.0152 \\
        \textbf{3} & 0.7273 & 0.0154  \\
        \textbf{4} & 0.7346 & 0.0154  \\
        \textbf{5}  & 0.7445 & 0.0155 \\
        \bottomrule
    \end{tabular}
    \label{bm25_performance}
\end{table}

%% file: tables/human_eval_rank_acc_259.tex
\begin{table}[htbp]
\centering
\caption{  Human Evaluation of all LLMs using just those 259 instructions that did not trigger a content filtering error. Due to Azure's content filtering system, 44 questions were unanswered for one or more of the GPT-4 model variants.}
\begin{tabular}{cccc}
\toprule
\textbf{Model} & \textbf{Context} & \% \textbf{Correct} $\uparrow$ & \textbf{Rank} $\downarrow$ \\
\midrule

GPT-4 (MR) & 32k & 68.3\% & 2.83 \\
GPT-4  & 32k & \textbf{69.9}\% & \textbf{2.42} \\
GPT-4 & 2048$^*$ & 52.5\% & 3.12 \\
Vicuña-13B & 2048 & 35.9\% & 4.01 \\
Vicuña-7B &	2048 & 34.4\% & 3.99 \\
MPT-7B-instruct 	& 2048 & 30.5\% & 4.61 \\
\bottomrule
\end{tabular}
\label{tab:metrics_correctness_and_rankings_no_content_filter}
\end{table}

%% file: tables/correct_303_by_label_updated.tex
\begin{adjustbox}{center}
\begin{tabular}{lcccccccc}

\toprule
 & \# & GPT-4  & GPT-4 & GPT-4 & Vicuña-13B  & Vicuña-7B & MPT-7B- & Macro \\

  &  & (32k + MR) & (32k) & (2k) & (2k) &  (2k) &  Instruct  (2k) & Average \\
 
Instruction Type &  &  &  &  &  &  &  &  \\
\midrule
Retrieve medical history & 67 & \textbf{0.701}& 0.627 & 0.552 & 0.358 & 0.343 & 0.313 & 0.483 \\
Retrieve imaging studies & 42 & \textbf{0.619} & 0.524 & 0.429 & 0.333 & 0.381 & 0.262 & 0.425 \\
Retrieve social history & 21 & 0.714 &  \underline{\textbf{0.762}} & 0.429 & 0.429 & 0.429 & 0.619 & 0.563 \\
Retrieve medication information & 22 & 0.545 & \textbf{0.591} & 0.455 & 0.409 & 0.318 & 0.455 & 0.462 \\
Retrieve laboratory studies & 12 & \textbf{0.667} & \textbf{0.667} & 0.333 & 0.333 & 0.250 & 0.333 & 0.431 \\
Provide patient education information & 7 & 0.571 & \textbf{0.714} & 0.571 & 0.143 & 0.143 & 0.143 & 0.381 \\
Retrieve appointment information & 13 & \textbf{0.538} & 0.462 & 0.385 & 0.231 & 0.231 & 0.308 & 0.359 \\
Retrieve demographics & 12 & \textbf{0.750} & 0.417 & 0.417 & 0.250 & 0.250 & 0.417 & 0.417 \\
Retrieve vitals & 10 & \textbf{0.700} & 0.600 & 0.300 & 0.200 & 0.100 & 0.500 & 0.400 \\
Retrieve genetic and family history & 6 &  \underline{\textbf{1.000}} & 0.833& 0.167 & 0.167 & 0.333 &  \underline{0.667} & 0.528 \\
Retrieve institutional policies & 6 & 0.667 & 0.667 &  \underline{\textbf{1.000}} & 0.500 & 0.333 & 0.167 & 0.556 \\
Retrieve provider referrals & 4 & 0.000 & 0.000 & \textbf{0.500} & 0.250 & 0.000 & 0.250 & 0.167 \\
Retrieve communications with patient & 1 & 0.000 & 0.000 &  \underline{\textbf{1.000}} &  \underline{\textbf{1.000}}& 0.000 & 0.000 & 0.333 \\
Calculate numerical measures & 6 & \textbf{0.667}& 0.500 & 0.500 & 0.333 & 0.167 & 0.167 & 0.389 \\
Provide risk assessment & 7 & 0.286 & \textbf{0.714} & 0.571 & 0.286 & 0.000 & 0.143 & 0.333 \\
Provide diagnosis assistance & 4 & \textbf{0.750} & 0.500 & \textbf{0.750} & 0.500 &  \underline{0.500} & 0.250 & 0.542 \\
Determine plan of care & 22 &  \textbf{0.909}& 0.773 & 0.636 & 0.364 & 0.455 & 0.227 & 0.561 \\
Other & 41 & 0.561 & 0.561 & \textbf{0.683} & 0.415 & 0.439 & 0.195 & 0.476 \\

\midrule
Macro Average &  & 0.591 & 0.551 & 0.538 & 0.361 & 0.260 & 0.301 & 0.434 \\
Micro Average &  & 0.650 & 0.601 & 0.518 & 0.350 & 0.333 & 0.317 & 0.461 \\
\midrule
\textbf{Total Number of Instructions} & 303 \\
\bottomrule
\end{tabular}
\end{adjustbox}
\label{tab:perfomrance_by_category}

%% file: tables/point_biserial_correct_rank.tex
\begin{table}[htbp]
\centering
\caption{  Point Biserial Correlation between Correctness and Human Ranking.}
\begin{tabular}{lcc}
\toprule
\textbf{Category} & \textbf{Avg. Corr} &  \textbf{95\% CI} \\
\midrule
Information Retrieval &  -0.78 & -0.80 to -0.76\\
Other & -0.74 & -0.79 to -0.69 \\
Care Planning  & -0.79 & -0.85 to -0.73 \\
Diagnosis Support & -0.77 & -0.87 to -0.67 \\
Calculation \& Scoring & -0.79 & -0.86 to -0.72 \\
\midrule
\textbf{Total} & -0.79 &  -0.86 to -0.72 \\
\bottomrule
\end{tabular}
\label{point_biseral}
\end{table}

%% file: tables/all_correct_instructions.tex
\begin{table*}[ht]
\caption{Instructions for which all model responses were deemed correct by a clinician. Comprises 22 (7.26\%) of the 303 instructions reviewed.}
\renewcommand{\arraystretch}{1.1} 
\begin{tabular}{R{11cm}R{6cm}}
\toprule
\textbf{Instruction} & \textbf{Subcategory} \\
\midrule
Provide a high-level summary of this patient's medical record. & Retrieve Medical History \\
Does this patient have any active chronic viral infections? & Retrieve Medical History \\
Has this patient has any adverse surgical outcomes & Retrieve Medical History \\
What was the mechanism of this patient's wrist trauma? & Retrieve Medical History \\
Was a decline in kidney function observed after iodinated contrast agent administration in this patient in the past? & Retrieve Medical History \\
Were any complications reported during this patient's last MRI exam? & Retrieve Imaging Studies \\
Are there any internal discrepencies between the findings section of the report and the impression section? & Retrieve Imaging Studies \\
Does this radiology report contain PHI? & Retrieve Imaging Studies \\
This patient has a singular lung nodule of currently 12 mm in the left lower lobe. Please provide information about the presence and size of this nodule from previous chest CT reports. & Retrieve Imaging Studies \\
Does this patient receive treatment for hyperthyroidism? & Retrieve Medication Information \\
Does she take anything at home for sleep? & Retrieve Medication Information \\
Are there any concerns for drug-drug interactions or potential contraindications for this patient? & Retrieve Medication Information \\
Has she ever reported not feeling safe at home? & Retrieve Social History \\
Are there any specific cultural or religious considerations that may impact the patient's healthcare decisions or treatment options? & Retrieve Social History \\
Does this patient have further appointments scheduled after his MRI examination in our hospital today? & Retrieve Appointment Information \\
What is the appropriate dose of Gadovist for this patient's planned MRI exam? Point me to our Wiki. & Retrieve Institutional Policies \\
Has the patient had a temperature spike in the last 24 hours? & Retrieve Vitals \\
What is the patient's ASCVD risk score & Calculate Numerical Measures \\
Based on my patient's current symptoms and medical history, do they need to be admitted to hospital? & Determine Plan of Care \\
This patient has a normal chest x-ray examination, draft a short report. & Other \\
We have implemented the production of a new PET tracer (Ga-FAPI). Draft an informative message to our referring providers at the medical oncology department & Other \\
Provide a summary of the current staging critieria for NSCLC and the respective source. & Other \\
\bottomrule
\label{tab:correct_instructions_across_all_models}
\end{tabular}
\end{table*}

%% file: tables/all_incorrect_instructions.tex
\begin{table*}[ht]
\caption{Instructions for which all model responses were deemed incorrect by a clinician. Comprises 33 (10.9\%) of the 303 instructions reviewed.}
\small
\renewcommand{\arraystretch}{1.1} 
\begin{adjustbox}{center}
\begin{tabular}{R{11cm}R{6cm}}
\toprule
\textbf{Instruction} & \textbf{Subcategory} \\
\midrule
Provide a summary of this patient's course in hospital. & Retrieve Medical History \\
Who is the resident doctor who last saw this patient? & Retrieve Medical History \\
Does the patient typically sleep a lot during the day? & Retrieve Medical History \\
This is a new patient, given their medical history, social history, and family history, please list the indicated guideline driven tests & Retrieve Medical History \\
Has the patient left the hospital against medical advice & Retrieve Medical History \\
When did this patient receive his last CT scan of the chest? & Retrieve Imaging Studies \\
Please summarize this patient's last abdominal CT report. & Retrieve Imaging Studies \\
These are MRI reports created as part of a clinical study. Remove all PHI from the reports and provide PHI-free versions that I can use in scientific presentation. & Retrieve Imaging Studies \\
This is a complex whole body PET/MR report. Provide a version of this report that is more concise and easier to read for the referring physician. & Retrieve Imaging Studies \\
Provide a list of diagnostic radiation dose exposure this patient had from CT and PET in the past. & Retrieve Imaging Studies \\
This is a report draft of this patient's pelvic MRI for staging of newly diagnosed rectal cancer. Is this report complete? What information should be added? & Retrieve Imaging Studies \\
which benzodiazepines has this patient used & Retrieve Medication Information \\
What is the patient's vaccination history for COVID & Retrieve Medication Information \\
Does this patient often ``No Show'' for scheduled appointments? & Retrieve Appointment Information \\
List all the providers who have been involved in my patient's care over the past year and for what reason. & Retrieve Appointment Information \\
How often has this patient ``No Showed'' for appointments in the past five years? & Retrieve Appointment Information \\
How long has this patient been in the hospital this stay? & Retrieve Appointment Information \\
Summarize all blood work the patient had in the past year. & Retrieve Laboratory Studies \\
What is the patient's last creatinine? & Retrieve Laboratory Studies \\
What is the phone number for the patient's emergency contact? & Retrieve Demographics \\
What was the total urine output over the past 24 hours? & Retrieve Vitals \\
Please show the patient's vital signs over the past 6 months & Retrieve Vitals \\
Create patient instructions that summarize all the patient's medications, including precautions on drug interactions, contraindications, and guidance on lifestyle precautions due to possible adverse side effects from medications & Provide Patient Education Information \\
Who referred the patient? & Retrieve Provider Referrals \\
Who was the referring provider and what is the consult question? & Retrieve Provider Referrals \\
This is a list of report drafts by a resident from today's chest x-ray examinations. Prioritize the order I should read them as an attending by clinical severity. & Provide Risk Assessment \\
Calculate the FRAX score for this patient. & Calculate Numerical Measures \\
What details are missing from this clinical history that may be important in developing a differential diagnosis? & Provide Diagnosis Assistance \\
Are any additional vaccines recommended for this patient? & Determine Plan of Care \\
Draft preconception counseling instructions individualized for this patient that incorporates pertinent prior medical history and medications & Other \\
Adjust d-dimer cutoffs by age to help rule out VTE & Other \\
What are the normal sizes of the liver, spleen, and kidneys for this pediatric patient? & Other \\
Given patient's labs, draft a message to the patient that is simple and clear notifiying them of their result and what it means & Other \\
\bottomrule
\end{tabular}
\end{adjustbox}
\label{tab:incorrect_instructions_across_all_models}
\end{table*}

%% file: tables/length_accuracy_303.tex
\begin{tabular}{lccrrrrrr}
\toprule
 & Token Length & Count & GPT-4 (32k + MR) & GPT-4 (32k) & GPT-4 (2k) & Vicuña-13B (2k) & Vicuña-7B (2k) & MPT-7B-Instruct (2k) \\
Quartiles &  &  &  &  &  &  &  &  \\
\midrule
1 & 9,390 - 39,076  & 79 & \textbf{0.671} & 0.608 & 0.532 & \underline{0.418} & 0.354 & \underline{0.354} \\
2 & 39,076 - 65,323  & 76 & \underline{\textbf{0.697}}& 0.605 & 0.500 & 0.289 & \underline{0.355} & 0.316 \\
3 & 65,323 - 114,850  & 75 & \textbf{0.640} & 0.573 & \underline{0.600} & 0.373 & 0.320 & 0.280 \\
4 & 114,850 - 496,148  & 73 & 0.589 & \underline{\textbf{0.616}} & 0.438 & 0.315 & 0.301 & 0.315 \\
\midrule
Total &  & 303 & \textbf{0.650} & 0.601 & 0.518 & 0.350 & 0.333 & 0.317 \\
\bottomrule
\end{tabular}

%% file: tables/length_accuracy_259.tex
\begin{tabular}{lcrrrrrrr}
\toprule
 & Token Length & Count & GPT-4 (32k + MR) & GPT-4 (32k) & GPT-4 (2k) & Vicuña-13B (2k) & Vicuña-7B (2k) & MPT-7B-Instruct (2k) \\
Quartiles &  &  &  &  &  &  &  &  \\
\midrule
1 & 9,390-40,139  & 68 & \underline{\textbf{0.721}} & \textbf{0.721} & 0.559 & \underline{0.412} & \underline{0.368} & \underline{0.338} \\
2 & 40,139-65,328  & 64 &\textbf{ 0.719} & 0.688 & 0.469 & 0.266 & 0.359 & 0.312 \\
3 & 65,328-114,779  & 65 & \textbf{0.662} & \textbf{0.662} & \underline{0.585} & 0.400 & 0.323 & 0.246 \\
4 & 114,779-496,148  & 62 & 0.629 & \underline{\textbf{0.726}} & 0.484 & 0.355 & 0.323 & 0.323 \\
\midrule
Total &  & 259 & 0.683 & \textbf{0.699} & 0.525 & 0.359 & 0.344 & 0.305 \\
\bottomrule
\end{tabular}

%% file: tables/average_rank_303.tex
\begin{tabular}{lcrrrrrrr}
\toprule
 & Token Length & Count & GPT-4 (32k + MR) & GPT-4 (32k) & GPT-4 (2k) & Vicuña-13B (2k) & Vicuña-7B (2k) & MPT-7B-Instruct (2k) \\
Quartiles &  &  &  &  &  &  &  &  \\
\midrule
1 & 9,390-39,076  & 79 & \textbf{2.772} & 2.829 & 3.203 &   \underline{3.772} &   \underline{3.829}  & 4.595 \\
2 & 39,076-65,323  & 76&   \underline{2.770} &   \underline{\textbf{2.579}} &   \underline{3.158} & 4.072 & 3.934 & 4.487 \\
3 & 65,323-114,850  & 75 &\textbf{ 2.833} & 2.953 & 2.773 & 3.867 & 3.993 & 4.580 \\
4 & 114,850-496,148  & 73 & 2.836 & \textbf{2.637} & 3.301 & 3.979 & 3.979 & 4.267 \\
\midrule
Total & & 303 & 2.802 &\textbf{ 2.751} & 3.109 & 3.921 & 3.932 & 4.485 \\
\bottomrule
\end{tabular}

%% file: tables/model_clinician_response_lengths.tex
\begin{tabular}{rrrrrrrr}
\toprule
  & GPT-4 (32k + MR) & GPT-4 (32k) & GPT-4 (2k) & Vicuña-13B (2k) & Vicuña-7B (2k) & MPT-7B-Instruct (2k) & Clinician\\
Percentile &  &  &  &  &  & & \\
\midrule
0 & 10.0 & 2.0 & 6.0 & 1.0 & 2.0 & 1.0 & 1.0 \\
25 & 96.5 & 20.0 & 18.0 & 22.5 & 19.0 & 11.0 & 20.0 \\
50 & 177.0 & 52.0 & 24.0 & 79.0 & 70.0 & 23.0 & 43.0 \\
75 & 256.0 & 100.0 & 66.5 & 182.0 & 158.5 & 231.0 & 90.0 \\
100 & 256.0 & 256.0 & 256.0 & 251.0 & 236.0 & 298.0 & 1148.0 \\
\bottomrule
\end{tabular}
\label{tab:Responses_rokne_length}

%% file: tables/comet_average_win_rate.tex
\begin{table}[tbp]
\centering
\caption{  Automatic Evaluation using COMET. Both the 303 question set (the full \dataset\ dataset) and the 259 question set (excluding instruction-EHR pairs that triggered Azure's content filter) are considered.}
\begin{tabular}{lccc}
\toprule
\textbf{Model} & \textbf{Context} &  \textbf{WR} $\uparrow$ & \textbf{WR} $\uparrow$  \\
& & \textbf{303} & \textbf{259} \\
\midrule

GPT-3.5  & 8192 & \textbf{0.61} & 0.60\\
LLaMA 2 13B Chat & 4096 & \textbf{0.61} & 0.60\\
GPT-4 (MR) &  32768$^\dag$ & 0.58 & 0.57\\
GPT-4  &  32768 & 0.56 & \textbf{0.64} \\
GPT-4 & 2048$^*$ & 0.56& 0.56 \\
LLaMA 2 7B Chat & 4096 & 0.51 & 0.50\\
Vicuña-7B &	2048 & 0.43 & 0.42 \\
Vicuña-13B & 2048 & 0.40 & 0.39 \\
MPT-7B-instruct 	& 2048 & 0.23 & 0.22\\
\bottomrule
\end{tabular}
\label{tab:comet_win_rate_post_evaluation}
\end{table}

%% file: tables/human_comet_win_rate.tex
\begin{table}[htbp]
\centering
\caption{  Human-evaluated Win Rate vs. COMET-evaluted  Win Rate using  303 instructions.}
\begin{tabular}{R{2cm}C{1.3cm}C{1.5cm}C{1.5cm}C{1.5cm}}
\toprule
\textbf{Model} & \textbf{Context} &  \textbf{WR} $\uparrow$ & \textbf{WR} $\uparrow$   & $|$\textbf{$\Delta$ WR}$|$ \\
& & \textbf{Human} & \textbf{COMET}  \\
\midrule
GPT-4 (MR)          &  32768$^\dag$  & 0.66          & \textbf{0.63} & 0.03 \\
GPT-4               & 32768          & \textbf{0.68} & 0.6           & 0.08\\
GPT-4               & 2048$^*$       & 0.6           & 0.6           & 0\\
Vicuña-13B          & 2048           & 0.40           & 0.45         & 0.05\\
Vicuña-7B           & 2048           & 0.40          & 0.48          & 0.08\\
MPT-7B 	& 2048           & 0.27          & 0.25          & 0.02 \\
\bottomrule
\end{tabular}
\label{tab:comet_human_win_comet}
\end{table}